\documentclass[Afour,sageh,times]{sagej}

\usepackage{amsmath}
\usepackage{amssymb}
\usepackage{graphicx}
\usepackage{booktabs}
\usepackage{tabularx}
\usepackage{url}
\usepackage{xltabular}
\usepackage{subcaption}
\usepackage{tikz}
\usetikzlibrary{calc}
\usepackage{multirow}

\usepackage[colorlinks,bookmarksopen,bookmarksnumbered,citecolor=red,urlcolor=red,linkcolor=red]{hyperref}
\hypersetup{
  pdftitle={Topometric Autonomous Vehicle Localization by Combining Visual Embeddings and Feed-Forward 3D Models},
  pdfauthor={Eulogio Quemada-Torres, Alberto Jaenal, Francisco-Angel Moreno, Javier Gonzalez-Jimenez},
  pdfkeywords={Visual localization, visual place recognition, topometric maps, Monte Carlo localization, feed-forward 3D models}
}
\usepackage{cleveref}

\usepackage{float}
\usepackage{placeins}

\floatstyle{ruled}
\newfloat{algorithm}{tbp}{loa}
\floatname{algorithm}{Algorithm}

\newenvironment{pseudocode}{%
  \begin{enumerate}
  \footnotesize
  \setlength{\itemsep}{1pt}
  \setlength{\parsep}{0pt}
  \setlength{\topsep}{2pt}
  \setlength{\partopsep}{0pt}
}{%
  \end{enumerate}
}

\newcommand{\pstate}{\item}
\newcommand{\pindent}{\hspace*{1em}}

\crefname{algorithm}{Algorithm}{Algorithms}
\Crefname{algorithm}{Algorithm}{Algorithms}

\begin{document}

%%%%%%%%%%%%%%%%%%%%%%%%%%%%%%%%%%%%%%%%%%%%%%%%%%%%%%%%%%%%%%%%%%%%%%%%%%%%%%%%
% Header information
%%%%%%%%%%%%%%%%%%%%%%%%%%%%%%%%%%%%%%%%%%%%%%%%%%%%%%%%%%%%%%%%%%%%%%%%%%%%%%%%

\runninghead{Quemada-Torres et al.}

\title{Topometric Autonomous Vehicle Localization by Combining Visual Embeddings and Feed-Forward 3D Models}

\author{Eulogio Quemada-Torres\affilnum{1}, Alberto Jaenal\affilnum{2}, Francisco-Angel Moreno\affilnum{1} and Javier Gonzalez-Jimenez\affilnum{1}}

\affiliation{
\affilnum{1}Machine Perception and Intelligent Robotics Group (MAPIR), Department of Systems Engineering and Automation, Malaga Institute for Mechatronics Engineering and Cyber-Physical Systems (IMECH.UMA), University of Malaga, Malaga, Spain\\
\affilnum{2}Robotics, Computer Vision and Artificial Intelligence Group (RoPeRT), Aragon Institute for Engineering Research (I3A), University of Zaragoza, Zaragoza, Spain
}

\corrauth{Eulogio Quemada-Torres, Machine Perception and Intelligent Robotics Group (MAPIR), University of Malaga, Bulevar Louis Pasteur 35, 29071 Malaga, Spain.}

\email{eulogioquemada@uma.es}

%%%%%%%%%%%%%%%%%%%%%%%%%%%%%%%%%%%%%%%%%%%%%%%%%%%%%%%%%%%%%%%%%%%%%%%%%%%%%%%%
% Abstract and keywords
%%%%%%%%%%%%%%%%%%%%%%%%%%%%%%%%%%%%%%%%%%%%%%%%%%%%%%%%%%%%%%%%%%%%%%%%%%%%%%%%

\begin{abstract}
% sections/00_abstract.tex

Effective Visual Localization (VL) requires a map of the environment that combines compactness for efficient scalability with robustness against visual appearance changes and metric precision. Through low-dimensional image embeddings, Visual Place Recognition (VPR) is able to successfully meet the first two requirements, but its low metric accuracy makes it less suitable than standard VL approaches based on local features or neural representations. This limitation can be overcome by integrating VPR with the accurate local trajectory estimates produced by feed-forward neural 3D geometry (FF3D) models. In this paper, we address sequential appearance-based localization through a topometric framework that iteratively combines probabilistic VPR with FF3D metric pose estimation in controlled image sets. Our approach proposes an automatic offline mapping tool that models the topometric pose-appearance interaction in the different parts of the scene. This map is later employed by an online particle filter that estimates the pose from odometry and belief over places for FF3D inference, successfully incorporating neural metric estimation into probabilistic appearance-based localization. We extensively evaluate the framework on three known benchmarks, demonstrating substantial improvements over existing appearance-based methods. The modularity of our approach allows the descriptor extractor and FF3D model to remain interchangeable, and a focused analysis further shows that sequential belief can mitigate severe failures under perceptual aliasing.

\end{abstract}


\maketitle

%%%%%%%%%%%%%%%%%%%%%%%%%%%%%%%%%%%%%%%%%%%%%%%%%%%%%%%%%%%%%%%%%%%%%%%%%%%%%%%%
% Main paper
%%%%%%%%%%%%%%%%%%%%%%%%%%%%%%%%%%%%%%%%%%%%%%%%%%%%%%%%%%%%%%%%%%%%%%%%%%%%%%%%

% sections/01_introduction.tex
% Longitud total entre intro + related (si hay): ~ 1-1.5 páginas

% Outline:
% --------
% Visual localization is important.
% VPR improved robustness.
% But retrieval is still coarse.
% Topometric PF methods help but scale poorly.
% Foundation models improve metric precision but are expensive.
% We combine both worlds efficiently.

\section{INTRODUCTION}
\label{sec:intro}

% 1. [1 párrafo] General problem. Long-term visual localization: why visual localization matters?
%   - GPS denied / unrealiable.
%   - Visual localization relies on cheap and ubiquitous sensors.
%   - But: appearance changes make localization difficult: illumination, weather, perceptual aliasing, etc.
%   - However: modern VPR descriptors improved robustness enormously (refs: NetVLAD, CosPlace, MixVPR, etc.)
%   - Although: retrieval alone is not enough for reliable metric localization.

% Reliable visual localization is a core requirement for autonomous robots and vehicles operating over previously mapped environments. Cameras provide a lightweight and widely available sensing modality, especially when GNSS is unreliable or unavailable, but long-term visual localization remains challenging due to illumination changes, weather, seasonal variations, viewpoint changes, and perceptual aliasing. Visual Place Recognition (VPR) addresses this problem by retrieving visually similar places from a database of georeferenced images, and modern global descriptors have substantially improved retrieval robustness under appearance changes~\cite{lowry2015visual,masone2021survey,arandjelovic2016netvlad,izquierdo2024salad}. However, retrieval alone still provides a coarse localization cue: it can identify candidate places or reference images, but it does not directly yield the metric pose accuracy required by downstream navigation tasks.

Navigating within a previously mapped environment is a fundamental capability for the autonomous operation of mobile robots and vehicles. Owing to the compactness, low cost, and widespread availability of cameras, visual localization~\citep{panek2026guide} has emerged as a reliable alternative to GNSS-based solutions, particularly in environments where satellite signals are unreliable or unavailable, e.g. indoor spaces or urban canyons.
In this context, local feature-based methods~\citep{schoenberger2016sfm} have reached remarkable maturity, increasingly integrating learning-based components across different stages of the localization pipeline~\citep{lindenberger2023lightglue} to efficiently deliver accurate pose estimates.
Nevertheless, long-term visual localization remains a challenge for them, due to significant changes in illumination, weather, and viewpoint. Furthermore, scaling these approaches to large environments remains difficult~\citep{toft2020long}, as the size of the map and both the associated computational and storage requirements increase substantially.

Visual Place Recognition (VPR)~\citep{lowry2015visual,masone2021survey} addresses these limitations by retrieving visually similar images from a georeferenced image database, using whole-image descriptors designed to remain robust under substantial appearance changes. However, this robustness generally comes at the expense of metric localization accuracy.
Consequently, VPR is commonly used in a coarse localization stage within hierarchical localization pipelines~\citep{sarlin2019coarse}.
Moreover, since VPR typically processes images independently, its performance may still be degraded by perceptual aliasing or severe environmental changes, leading to incorrect place matches. %it fails to yield the metric pose accuracy required by navigation tasks, which

% 2. [1 párrafo] Topometric PF methods help but still limited (refs: Alberto, etc.)
%   - Limited metric precision inside a place
%   - Scalability issues
%   - Redundancy in dense maps
%   - Manual tuning of map building process
%   - (Alberto IJRR) Our previous work addressed these challenges by 
%       - Gaussian cluster-based PF localization
%       - Benefits: Compact representation, probabilistic reasoning, robustness
%       - Limitations: Predefined number of clusters & sensitive clustering initialization, no metric refinement within clusters

Sequential probabilistic localization~\citep{blanco2008toward} alleviates these limitations by exploiting the temporal coherence naturally available in robot trajectories, combining successive observations through Bayesian filtering techniques.
Unlike most sequential localization approaches, which rely on local visual features, this work focuses exclusively on methods built upon VPR descriptors~\citep{xu2020probabilistic}, as they provide a more compact and appearance-robust representation of the environment.
%
%evertheless, the localization accuracy achieved by these methods remains fundamentally limited by the spatial resolution of the underlying topological representation of the scene and the relatively low precision of VPR-based continuous pose estimation, resulting in coarse localization compared with local feature-based approaches.
Nevertheless, these methods remain fundamentally limited in their accuracy. This is due to the coarse spatial resolution of such representations, combined with the low precision of VPR-based continuous pose estimation, leaving them significantly less accurate than local feature-based approaches.

Recent feed-forward neural 3D geometry (FF3D) models, such as DUSt3R~\citep{wang2024dust3r} and VGGT~\citep{wang2025vggt}, can address this limitation by directly inferring accurate geometric information from sparse image sets.
Despite their impressive performance, their computational cost increases with the number of input images, making their direct application to dense georeferenced image maps computationally inefficient and highly dependent on the selection of informative reference images.
Therefore, a practical sequential localization framework integrating FF3D models must efficiently employ a small set of plausible map images for neural metric inference, thereby enabling scalability while maintaining localization accuracy.
%Topometric probabilistic localization mitigates this limitation by combining visual observations with temporal filtering and a structured map representation. In this context, appearance-map abstraction reduces the redundancy of dense image databases and enables localization over compact topological places instead of individual reference images. Nevertheless, once the belief has converged to the correct place, the metric precision remains limited by the spatial extent and internal variability of that place. Recent feed-forward neural geometry models provide a complementary opportunity for metric refinement. Methods such as DUSt3R, MASt3R, VGGT, and Depth Anything 3 can infer geometric structure or camera poses from sparse image sets~\cite{wang2024dust3r,leroy2024mast3r,wang2025vggt,lin2025depthanything3}. These models are attractive for refining a coarse topometric estimate, but applying them directly to a dense reference map is computationally expensive and sensitive to the choice of input images. Redundant, ambiguous, or geometrically inconsistent references may degrade the estimate. Therefore, a practical localization system must restrict neural metric inference to a small set of plausible and informative map images.

% 3. [1 párrafo] Connection to Jaenal et al. after the general motivation.
%   - Appearance-map abstraction and recursive topometric localization are the inherited backbone.
%   - Limitations: predefined/selected number of clusters, sensitive clustering initialization, no metric refinement within clusters.
To address these limitations, we propose a topometric localization framework that combines the robustness of sequential VPR-based localization with the metric accuracy of feed-forward neural 3D geometry (FF3D) models. The proposed framework consists of two complementary stages.
The first stage is an offline map construction process inspired by~\cite{jaenal2022unsupervised}, which uses VPR embeddings to generate probabilistic topological places, but is
% the original abstraction procedure relies on an iterative clustering algorithm that 
highly sensitive to initialization, making it inefficient for large-scale environments. 
%and requires multiple restarts, making it computationally inefficient for large-scale environments. 
We overcome this limitation by employing HDBSCAN~\citep{campello2013hdbscan} to automatically discover topological places without requiring a predefined number of clusters.
The second stage is an online topometric particle filter built upon~\cite{jaenal2023sequential}, which provides robust topological tracking using VPR descriptors. We overcome the limited metric localization accuracy from the original work by introducing a cluster-conditioned neural metric refinement stage that uses probabilistic belief to query a FF3D model with map images. This significantly improves localization accuracy while preserving the robustness and scalability of the underlying topometric framework.

The contributions of this work are fourfold:
(i) a scalable topometric probabilistic localization framework that combines sequential VPR-based localization with neural metric pose estimation through a modular architecture, treating both visual embedding extraction and neural metric inference as plug-and-play components, thereby remaining agnostic to the underlying network architectures;
(ii) a probabilistic place construction strategy that automatically discovers topological places from map features without requiring a predefined number of clusters;
(iii) a representative-image selection and cluster-conditioning mechanism that enables the scalable integration of FF3D models into probabilistic localization over topometric appearance maps% by restricting neural metric inference to a compact set of informative map images
; and
(iv) an extensive experimental validation across multiple environments and operating conditions, assessing localization accuracy, scalability, and the effectiveness of the proposed framework in comparison with the state of the art.

\section{RELATED WORK}
\label{sec:related}

\subsection{Visual Place Recognition}
\label{sec:related:vpr}

Visual Place Recognition (VPR) addresses the problem of retrieving images similar to a query from a database of georeferenced images even under severe appearance changes~\citep{lowry2015visual,masone2021survey}. Consequently, visual localization, loop closure, and hierarchical localization pipelines~\citep{sarlin2019coarse,toft2020long,arnold2022map,dong2025reloc3r} have widely adopted VPR to provide a coarse localization prior, as it relies on compact global descriptors that allow efficient and appearance-robust querying over large image databases without requiring explicit geometric verification. This scalability has made VPR one of the most widely adopted front-ends for long-term visual localization and large-scale mapping.

Pioneering learned global descriptors such as NetVLAD~\citep{arandjelovic2016netvlad} established trainable aggregation as a strong paradigm for place recognition. More recent descriptors have improved accuracy and scalability through classification-based training~\citep{berton2022cosplace}, feature mixing~\citep{alibey2023mixvpr}, viewpoint-aware training~\citep{berton2023eigenplaces}, optimal-transport aggregation~\citep{izquierdo2024salad}, and learnable query-based aggregation~\citep{ali2024boq}. These methods provide increasingly robust appearance cues under viewpoint, illumination, and seasonal changes. Nevertheless, VPR remains fundamentally an image retrieval mechanism. It can identify visually plausible places or reference images, but it does not by itself provide the precise metric pose estimates required by downstream navigation tasks. Some approaches have explored pose interpolation~\citep{pion2020benchmarking} or joint learning of VPR and pose information~\citep{thoma2020geometrically}, although their metric accuracy remains far from dedicated pose-estimation methods. Consequently, while VPR scales efficiently to very large image collections, improving its metric localization accuracy remains an active research challenge.

Another major limitation of VPR is that nearest-neighbor retrieval in descriptor space can fail under perceptual aliasing when spatially distant places exhibit similar visual structures. Several works have therefore modeled retrieval uncertainty or estimated the reliability of individual predictions~\citep{warburg2021bayesian,cai2022stun,zaffar2024estimation,miller2026through}. However, these approaches primarily characterize uncertainty at the retrieval level. Sequential probabilistic localization addresses these limitations by maintaining multiple competing place hypotheses and propagating their uncertainty over time.

\subsection{Topometric Appearance-based Localization}
\label{sec:related:topometric}

Topometric localization combines appearance cues with structured map representations and temporal reasoning, allowing a system to maintain multiple place hypotheses while retaining metric information for navigation~\citep{blanco2008toward}. Classical appearance-based localization introduced several complementary ideas. FAB-MAP proposed a probabilistic formulation that explicitly accounts for perceptual aliasing~\citep{cummins_fab-map_2008}, whereas SeqSLAM exploits consistency across image sequences to recognize previously traversed routes under severe appearance changes~\citep{milford2012seqslam}. Hybrid metric-topological maps combine local metric information with higher-level place structure~\citep{blanco2008toward}, while particle filters provide a natural framework for recursive multi-hypothesis localization~\citep{dellaert1999monte}.

More recently, VPR has been incorporated into probabilistic localization frameworks that operate exclusively on global appearance descriptors, avoiding the need for local feature extraction and geometric verification. These methods preserve the scalability and robustness of VPR while exploiting temporal consistency to improve localization reliability. ~\cite{xu2020probabilistic} reformulate VPR retrieval as a probabilistic observation for hierarchical localization, providing a more informative spatial prior for subsequent metric pose estimation. Their later topometric formulation~\citep{xu2021probabilistic} extends this idea by incorporating full three-degree-of-freedom odometry and an explicit off-map state to handle route deviations. PlaceNav~\citep{suomela2024placenav} instead employs VPR descriptors within a Bayesian navigation framework, using probabilistic filtering to improve temporal consistency during topological navigation. These works demonstrate that probabilistic reasoning significantly improves the robustness of VPR-based localization while preserving its scalability. However, they remain fundamentally limited by the metric precision achievable from global descriptors alone, and they neither address the automatic abstraction of dense georeferenced image maps nor exploit recent FF3D models for metric pose refinement.

The closest predecessors of our work are the appearance-map abstraction proposed in~\cite{jaenal2022unsupervised} and its sequential localization extension~\citep{jaenal2023sequential}. The former compresses a dense database of georeferenced images into probabilistic places characterized by pose and appearance distributions, reducing map redundancy and representing uncertainty at the place level. The latter incorporates this representation into a sequential topometric filter and introduces place-specific appearance--pose models for continuous localization. These works provide the probabilistic map and filtering backbone adopted here. However, their map construction requires an offline procedure to determine the number of places and remains sensitive to clustering initialization, while their metric accuracy is bounded by the spatial resolution of the local appearance--pose model. Our work addresses these limitations by introducing automatic place discovery together with belief-guided FF3D metric refinement.

\subsection{Feed-forward Geometry Models}
\label{sec:related:ff_geometry}

Classical visual localization often estimates camera pose by matching local image features against a 3D map or an image database with geometric verification. Structure-from-Motion and local-feature pipelines remain among the most established solutions for accurate visual localization~\citep{schoenberger2016sfm,lindenberger2023lightglue}. However, they typically require explicit feature matching, map reconstruction, or geometric optimization, which can be expensive to maintain in large-scale or long-term localization scenarios.

Recent feed-forward geometry models provide a complementary source of metric information. Methods such as DUSt3R~\citep{wang2024dust3r}, VGGT~\citep{wang2025vggt}, and Depth Anything 3~\citep{lin2025depthanything3} can infer 3D structure and camera poses from sparse sets of input images. When applied to map-based localization, however, these models present several practical limitations: (i) their computational cost generally increases with the number of input views, (ii) their pose estimates depend strongly on the selected references, and (iii) the predicted geometry must be aligned with the georeferenced map frame. Dense image maps exacerbate these difficulties by introducing redundant and potentially ambiguous references, making straightforward FF3D inference difficult to scale to large environments. Our approach addresses these issues by exploiting the probabilistic topometric belief to restrict FF3D inference to a compact and informative subset of representative map images, thereby preserving scalability while achieving accurate metric localization.

\section{SYSTEM DESCRIPTION}
\label{sec:description}

\begin{figure*}[t]
    \centering
    \resizebox{\textwidth}{!}{%
        \begin{tikzpicture}[x=1bp, y=1bp]

            % PDF base:
            % offline map building + hybrid initialization
            % + marco completo de online localization.
            \node[
                anchor=south west,
                inner sep=0
            ] at (0,0) {%
                \includegraphics[
                    width=900.9bp,
                    height=540.4bp
                ]{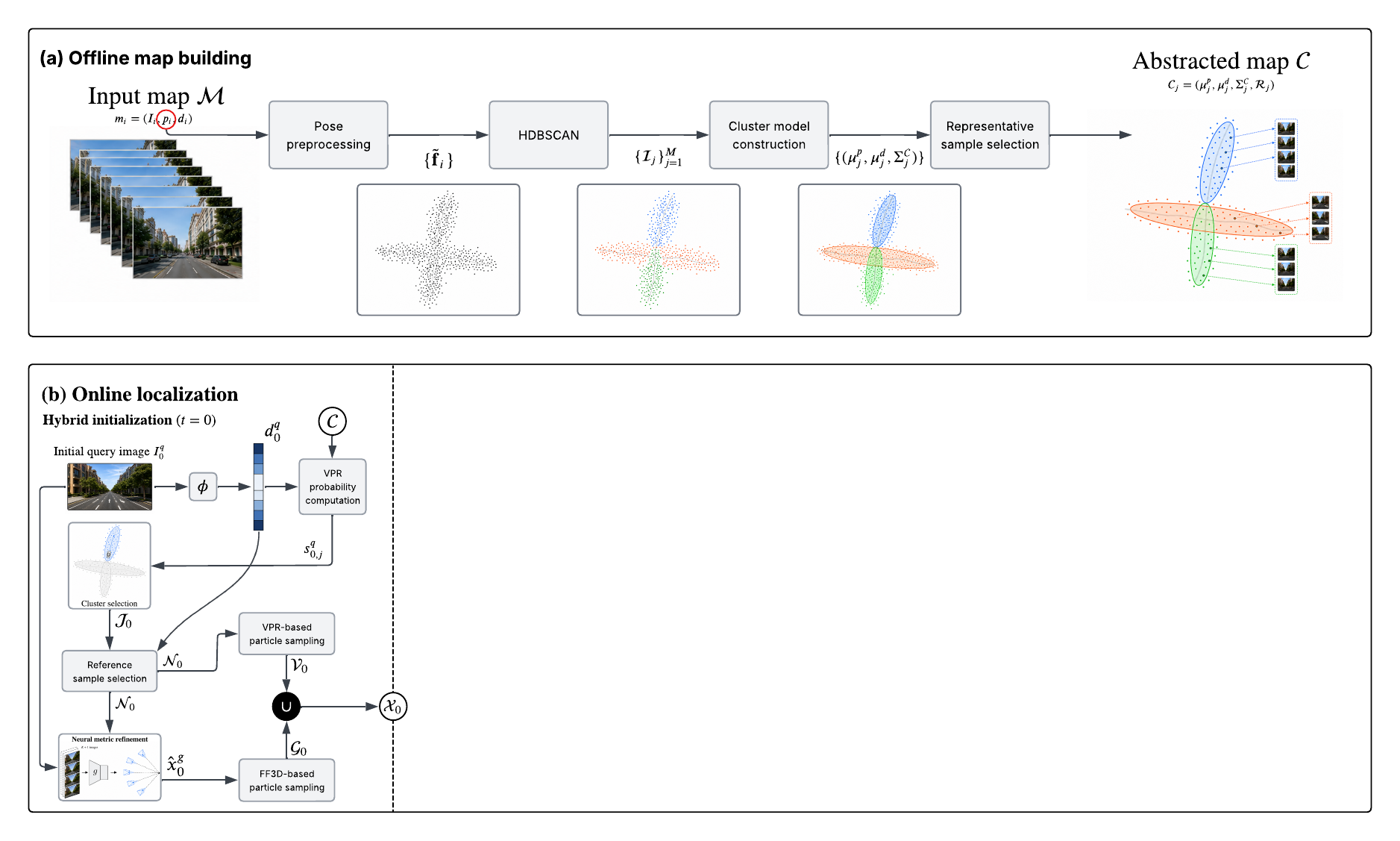}%
            };

            % Normal operation:
            % se elimina el espacio vacío de su PDF y se coloca
            % exactamente en el hueco derecho de base.pdf.
            \node[
                anchor=south west,
                inner sep=0
            ] at (262.3,19.75) {%
                \includegraphics[
                    trim=300.5bp 088.65bp 205bp 75.6bp,
                    clip,
                    width=615.15bp,
                    height=285.75bp
                ]{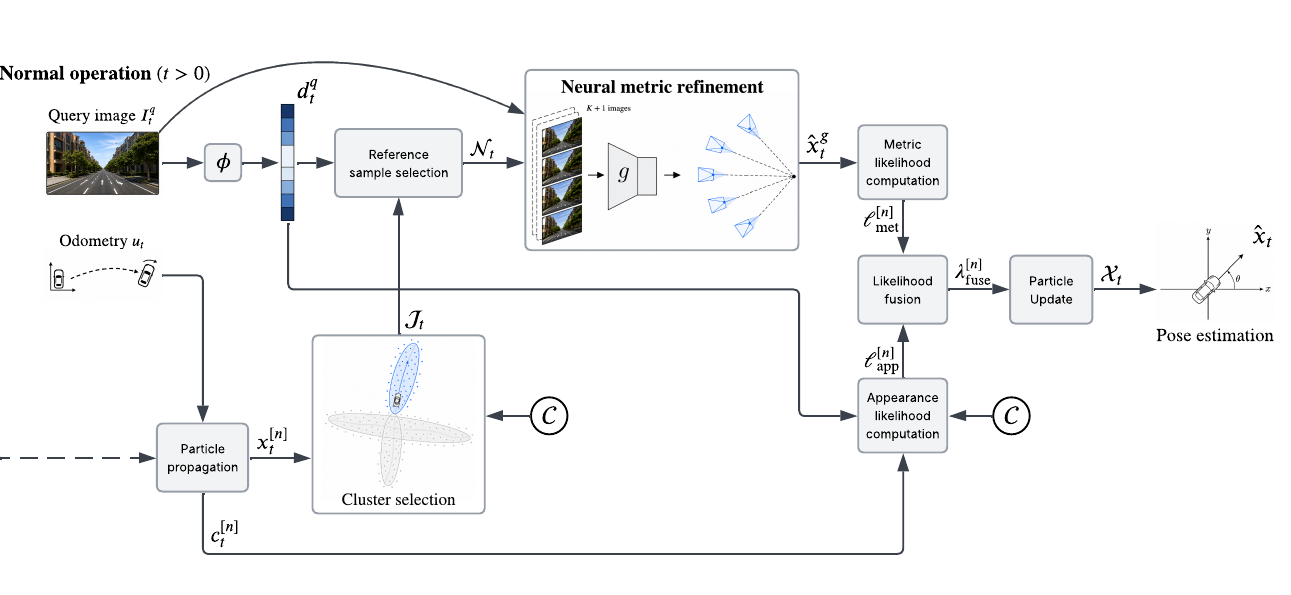}%
            };

        \end{tikzpicture}%
    }
    \vspace{-22px}

    \caption{
        Overview of the proposed topometric localization pipeline. The offline stage builds an abstracted appearance map from georeferenced images by constructing topological places and selecting representative samples. The online stage uses the resulting map in a topometric particle filter that combines appearance-based weighting with cluster-conditioned neural pose estimation.
    }
    \vspace{-10px}
    \label{fig:pipeline}
\end{figure*}

\Cref{fig:pipeline} provides an overview of the proposed two-stage topometric localization framework. The offline stage constructs a compact topometric appearance map from a dense set of georeferenced images by automatically discovering topological places and selecting representative map samples. The online stage estimates the camera pose through a topometric particle filter that combines odometry, VPR descriptors, and cluster-conditioned neural metric observations. Compared with~\cite{jaenal2023sequential}, our method introduces automatic map abstraction, hybrid particle initialization, and FF3D-based metric pose refinement.

\paragraph{Naming convention.}
Throughout the paper, indices $i$, $j$, and $[n]$ refer to map samples, clusters, and particles, respectively. We denote by $\phi$ the global image descriptor extractor (e.g., NetVLAD), which maps an image $I$ to a descriptor $d=\phi(I)\in\mathbb{R}^{D}$. Similarly, the feed-forward geometry model $g$ (e.g., VGGT) maps a set of input images to the corresponding set of camera transforms as
\begin{equation}
  \left\{ \hat{T}_i \right\}_{i=1}^{N}
  =
  g\!\left( \left\{ I_i \right\}_{i=1}^{N}\right),
  \label{eq:ff3d}
\end{equation}
where each transform $\hat{T}_i\in\mathrm{SE}(3)$ represents the pose of the corresponding input image in the internal coordinate frame estimated by the model. Both $\phi$ and $g$ are treated as interchangeable black-box components, allowing the proposed framework to operate with different descriptor extractors and FF3D models without modifying the underlying probabilistic formulation.

The remainder of this section first describes the offline map construction process and subsequently the online localization algorithm.

\subsection{Offline Map Building}
\label{sec:description:map}

The offline map building stage, illustrated in \Cref{fig:pipeline}(a), constructs a compact topometric representation \(\mathcal{C}\) from a dense georeferenced image map \(\mathcal{M}\), which contains data collected during one or more reference traversals. The resulting representation is subsequently employed during online localization.

Formally, let the input map be
\begin{equation}
  \mathcal{M} = \{m_i\}_{i=1}^{N},
  \qquad
  m_i = (I_i,\, p_i,\, d_i),
  \label{eq:map}
\end{equation}
where \(I_i\) is an image, \(p_i=(x_i,y_i,\theta_i)\in \mathrm{SE}(2)\) is its 2D pose within the map, and \(d_i\) is its associated global descriptor. The global descriptors serve two purposes: they characterize the appearance distribution associated with each topological place and, through cosine distance in descriptor space, guide the selection of visually diverse representative samples.

The output of the offline stage is
\begin{equation}
  \mathcal{C} = \{C_j\}_{j=1}^{M},
  \qquad
  C_j =
  \left(
    \mu_j^p,\,
    \mu_j^d,\,
    \Sigma_j^{\mathcal{C}},\,
    \mathcal{R}_j
  \right),
  \label{eq:abstracted_map}
\end{equation}
which defines the abstracted topometric map, where \(\mu_j^p\), \(\mu_j^d\), and \(\Sigma_j^{\mathcal{C}}\) correspond to the mean pose, mean descriptor, and joint pose--descriptor covariance associated with the \(j\)-th topological place, respectively, computed as described in \Cref{sec:description:cluster_model}. Finally, \(\mathcal{R}_j \subseteq \mathcal{M}\) denotes the representative sample set selected according to the procedure described in \Cref{sec:description:representatives}, which is retained for subsequent FF3D inference during online localization.

\subsubsection{Pose preprocessing and HDBSCAN.}
\label{sec:description:map:hdbscan}

We employ HDBSCAN~\citep{campello2013hdbscan} to construct the topological partition of the map. This density-based clustering algorithm automatically estimates the number of clusters from the data, avoiding the need to predefine a fixed number of topological places. Clustering is performed on the planar map poses by converting each pose \(p_i=(x_i,y_i,\theta_i)\) into the following pose-feature vector:
\begin{equation}
  \mathbf{f}_i
  =
  \bigl[x_i,\, y_i,\, \sin\theta_i,\, \cos\theta_i\bigr]^\top
  \in \mathbb{R}^{4}.
  \label{eq:pose_features}
\end{equation}

The sine and cosine components encode the heading angle without introducing the artificial discontinuity at the \( -\pi/\pi \) boundary. Since HDBSCAN relies on Euclidean distances in this feature space, the pose-feature vectors are standardized before clustering:
\begin{equation}
  \tilde{\mathbf{f}}_i =
  \frac{\mathbf{f}_i-\boldsymbol{\mu}_f}{\boldsymbol{\sigma}_f},
  \label{eq:pose_feature_standardization}
\end{equation}
where \(\boldsymbol{\mu}_f\) and \(\boldsymbol{\sigma}_f\) are the mean and standard deviation of the pose-feature vectors over the map. This normalization makes the translational and angular components comparable in the distance computation, preventing the translational coordinates from dominating solely because of their numerical scale.

HDBSCAN is then applied to the set \(\{\tilde{\mathbf{f}}_i\}\), yielding the partition \(\{\mathcal{I}_j\}_{j=1}^{M}\) of map samples, where
\(\mathcal{M}_j=\{m_i\}_{i\in\mathcal{I}_j}\subseteq\mathcal{M}\)
denotes the subset of map samples assigned to the \(j\)-th topological place. Samples initially labeled as noise by HDBSCAN are reassigned to the nearest non-noise cluster centroid in the standardized pose-feature space before the place statistics are computed, ensuring that every map sample contributes to a topological place.

\subsubsection{Cluster model construction.}
\label{sec:description:cluster_model}

For each topological place \(C_j\), the pose mean \(\mu_j^p \in \mathrm{SE}(2)\) is computed from the poses of the map samples in \(\mathcal{M}_j\), while the descriptor mean is given by
\begin{equation}
  \mu_j^d =
  \frac{1}{|\mathcal{M}_j|}
  \sum_{m_i \in \mathcal{M}_j} d_i .
  \label{eq:cluster_means}
\end{equation}

The corresponding joint pose--descriptor covariance, \(\Sigma_j^{\mathcal{C}}\), is approximated by the compact block-diagonal representation proposed in~\cite{jaenal2022unsupervised}:
\begin{equation}
  \Sigma_j^{\mathcal{C}} =
  \begin{bmatrix}
    \Sigma_j^p & 0 \\
    0 & \sigma_{j,d}^{2}
  \end{bmatrix}
  \in \mathbb{R}^{4\times4},
  \label{eq:cluster_covariance}
\end{equation}
where the zero off-diagonal blocks reflect the assumption of conditional independence between pose and appearance. The matrix \(\Sigma_j^p\in\mathbb{R}^{3\times3}\) models the pose uncertainty of the topological place and is estimated from the \(\mathrm{SE}(2)\) tangent-space deviations of the map sample poses with respect to \(\mu_j^p\), preserving the correlations between translation and heading. Descriptor variability is summarized by the scalar variance \(\sigma_{j,d}^{2}\), computed from the squared descriptor distances to \(\mu_j^d\). This scalar parameterizes the isotropic descriptor covariance \(\Sigma_j^d=\sigma_{j,d}^{2}I_D\), avoiding the need to estimate and store a full \(D\times D\) covariance matrix. Further implementation details can be found in~\cite{jaenal2022unsupervised}.

\subsubsection{Representative sample selection.}
\label{sec:description:representatives}

Although the pose and descriptor distributions of each topological place are summarized by the cluster model, online neural metric refinement requires access to map images. Retaining every image within a topological place would, however, produce a large deployable map with substantial visual redundancy. Each map subset \(\mathcal{M}_j\) is therefore reduced to a representative sample set \(\mathcal{R}_j\subseteq\mathcal{M}_j\) using farthest-point sampling (FPS) in descriptor space. This abstraction reduces image storage while preserving visually diverse candidates for online reference selection.

The FPS procedure preserves a minimum of two representative images while iteratively selecting samples that maximize the visual diversity of the selected set. The process terminates once the remaining descriptor diversity falls below the threshold \(\tau_{\mathrm{diversity}}\). The complete selection procedure is summarized in \Cref{alg:representative_selection}.

It is worth emphasizing that all samples in \(\mathcal{M}_j\) contribute to the probabilistic cluster model, whereas only the selected representatives are retained as candidates for online reference selection. Before map serialization, these images are resized to the input resolution required by \(g\) and JPEG-encoded, further reducing the deployable map size.

\begin{algorithm}[!htbp]
    \caption{Per-cluster representative sample selection}
    \label{alg:representative_selection}
    \textbf{Input:} cluster sample set $\mathcal{M}_j$ and diversity threshold $\tau_{\mathrm{diversity}}$.\\
    \textbf{Output:} representative sample set $\mathcal{R}_j\subseteq\mathcal{M}_j$.
    \begin{pseudocode}
        \pstate L2-normalize the descriptor of every sample in $\mathcal{M}_j$.
        \pstate Compute the mean $\bar d_j$ of the normalized descriptors and L2-normalize it.
        \pstate Select the sample whose descriptor has the highest cosine similarity to $\bar d_j$ and initialize $\mathcal{R}_j$ with it.
        \pstate \textbf{while} unselected samples remain \textbf{do}
        \pstate \pindent For each unselected sample $m_i$, compute
        \[
        \delta_i=\min_{m_r\in\mathcal{R}_j}d_{\cos}(d_i,d_r).
        \]
        \pstate \pindent \textbf{if} $|\mathcal{R}_j|\geq2$ and $\max_i\delta_i<\tau_{\mathrm{diversity}}$ \textbf{then break}.
        \pstate \pindent Select $m_{i^*}$, where $i^*=\operatorname*{arg\,max}_i\delta_i$, and add it to $\mathcal{R}_j$.
        \pstate \textbf{end while}
        \pstate \textbf{return} $\mathcal{R}_j$.
    \end{pseudocode}
\end{algorithm}

% =========== ONLINE LOCALIZATION ================
\subsection{Online Localization}
\label{sec:description:localization}

Once the topometric map has been built, online localization is performed using a topometric particle filter that combines probabilistic place tracking with cluster-conditioned neural metric refinement (see \Cref{fig:pipeline}(b) for an overview). At each time step \(t\), the filter receives a query image \(I_t^q\) and an odometry measurement \(u_t\in\mathrm{SE}(2)\), computes the descriptor \(d_t^q=\phi(I_t^q)\), and computes the output camera pose \(\hat{x}_t\).
The posterior belief is represented by the particle set
\begin{equation}
  \mathcal{X}_t =
  \left\{
    \left(
      x_t^{[n]},\,
      c_t^{[n]},\,
      w_t^{[n]}
    \right)
  \right\}_{n=1}^{N_p},
  \label{eq:particle_state}
\end{equation}
where \(x_t^{[n]}\in\mathrm{SE}(2)\) denotes the particle pose, \(c_t^{[n]}\) the associated topological place, and \(w_t^{[n]}\) its normalized importance weight.

The localization process is initialized with a hybrid mechanism that models initial spatial uncertainty by combining FF3D pose estimates with multi-hypothesis appearance anchors. Once initialized, each step comprises odometry-driven particle propagation; cluster-conditioned neural metric refinement, where representative map images from high-belief places are processed by the FF3D model to yield an aligned metric pose observation; and a dual-likelihood particle update that weights particles by appearance and metric scores before pose extraction and conditional resampling.

\subsubsection{Hybrid Particle Initialization.}
\label{sec:description:particle_initialization}

The proposed hybrid initialization operates over an initial set of candidate places \(\mathcal{J}_0\), which holds multiple appearance-based pose hypotheses derived from VPR. To obtain such a set, the system computes the VPR probabilities \(s_{0,j}^q=P(d_0^q\mid C_j)\) of all clusters using the initial query descriptor $d_0^q$ and the descriptor distribution parameters previously described in \Cref{sec:description:cluster_model}. 
Then, \(\mathcal{J}_0\) is obtained using the acceptance ratio 
\begin{equation}
  \mathcal{J}_0 =
  \left\{
    j : s_{0,j}^q \geq
    \tau_{\mathrm{vpr}}\, s_{0,j^\star}^q
  \right\},
  \label{eq:init_vpr_regions}
\end{equation}
where $j^\star=\operatorname*{arg\,max}_j s_{0,j}^q$ denotes the most probable map cluster, and $\tau_{\mathrm{vpr}}$ is a fixed relative threshold.
Applying the reference sample selection and metric pose estimation procedures detailed in \Cref{sec:description:neural_metric} to $\mathcal{J}_0$, we first obtain a set of selected representative samples $\mathcal{N}_0$ and an initial pose estimate $\hat{x}_0^g$ through a FF3D model.

The initial particle set is then divided into two complementary groups: $\mathcal{V}_0$ and $\mathcal{G}_0$. The VPR-based group $\mathcal{V}_0$ is formed by $\beta N_p$ particles, which preserve multiple hypotheses by using the known map poses of the selected representative samples. The remaining $(1-\beta)N_p$ particles form the FF3D-based group $\mathcal{G}_0$, which represents the single metric hypothesis $\hat{x}_0^g$.

$\mathcal{V}_0$ is initialized by a weighted assignment to each map representative, computed through an appearance-based normalized score between the query and sample $m_a \in\mathcal{N}_0$:
\begin{equation}
    \pi_a =
    \frac{
    \left[\left\langle \mathbf{d}_a,\mathbf{d}_0^q\right\rangle\right]_+
    }{
    \sum_{m_b\in\mathcal{N}_0}
    \left[\left\langle \mathbf{d}_b,\mathbf{d}_0^q\right\rangle\right]_+
    },
    \label{eq:vpr_particle_assignment}
\end{equation}
where $\langle\cdot,\cdot\rangle$ denotes cosine similarity and $[z]_+=\max(0,z)$; uniform weights are used if the denominator is zero. This score is then used to distribute particles among the selected samples, following $ N_a \approx \pi_a\,|\mathcal{V}_0|$ so that more particles are assigned to references with higher similarity.
% \begin{equation}
%   N_a \approx \pi_a\,|\mathcal{V}_0|
%   %,
%   %\qquad
%   %\sum_{m_a\in\mathcal{N}_0} N_a = |\mathcal{V}_0|,
%   \label{eq:init_reference_counts}
% \end{equation}
For each selected sample $m_a\in\mathcal{N}_0$, its $N_a$ assigned particles are sampled around the corresponding known map pose $p_a$ using Gaussian perturbations in $\mathrm{SE}(2)$. 

In turn, particles in $\mathcal{G}_0$ are sampled around the FF3D estimate $\hat{x}_0^g$ with noise $\xi$ in the tangent space of $\mathrm{SE}(2)$.

Finally, all particles in both groups are initially assigned uniform weights,
\begin{equation}
  w_0^{[n]} = \frac{1}{N_p},
  \qquad n=1,\ldots,N_p.
  \label{eq:init_weights}
\end{equation}

\subsubsection{Particle Propagation and Cluster Selection.}
\label{sec:description:propagation_and_cluster}

After initialization, the particle filter propagates each particle through odometry:
\begin{equation}
\begin{aligned}
  {x}_t^{[n]} &=
  x_{t-1}^{[n]} \oplus u_t \oplus \epsilon_t^{[n]},\\
  \epsilon_t^{[n]} &= \exp(\eta_t^{[n]}),\quad
  \eta_t^{[n]} \sim \mathcal{N}(0,Q),\\
\end{aligned}
  \label{eq:particle_prediction}
\end{equation}
where $\oplus$ denotes $\mathrm{SE}(2)$ composition, $\eta_t^{[n]}\in\mathbb{R}^3$ is a sampled perturbation in the tangent space of $\mathrm{SE}(2)$, and $Q\in\mathbb{R}^{3\times3}$ is the motion-noise covariance.

Subsequently, each particle is associated with the cluster that maximizes its pose likelihood:
\begin{equation}
  {c}_t^{[n]} =
  \operatorname*{arg\,max}_{j}
  P\!\left(x_t^{[n]} \mid C_j\right).
  \label{eq:particle_cluster_assignment}
\end{equation}
We then use the previous step weights of the propagated particles to induce a belief over topological places:
\begin{equation}
  b_t(j) =
  \sum_{n:\, c_t^{[n]} = j} w_{t-1}^{[n]} ,
  \label{eq:cluster_belief}
\end{equation}
Finally, the set of candidate clusters is selected through a belief ratio criterion,
\begin{equation}
  \mathcal{J}_t =
  \left\{
    j \;:\; b_t(j) \geq \tau_{\mathrm{cl}}\, b_t(j_t^\star)
  \right\},
  \label{eq:selected_clusters}
\end{equation}
where $j_t^\star=\operatorname*{arg\,max}_j b_t(j)$ is the dominant cluster and $\tau_{\mathrm{cl}}$ is a fixed relative threshold.

\subsubsection{Neural Metric Refinement.}
\label{sec:description:neural_metric}

At each time step $t$, given the candidate place set $\mathcal{J}_t$ obtained from the particle belief (or from VPR for $\mathcal{J}_0$), the system computes an aligned metric pose observation $\hat{x}_t^g \in \mathrm{SE}(2)$ through two sequential steps: (i) reference sample selection to determine a subset of representative map samples as input, and (ii) feed-forward metric pose estimation.

\paragraph{Reference sample selection.}
\label{sec:description:reference_selection}

%At each time step, the particle filter belief restricts neural metric refinement to a subset of likely topological places.
%
%Specifically, the candidate place set \(\mathcal{J}_t\) determines the representative map samples that are eligible for reference selection. At initialization, \(\mathcal{J}_0\) is obtained from the VPR-based place hypotheses, whereas during regular localization it is derived from the current particle belief. This belief-guided selection substantially reduces the computational cost of FF3D inference while preserving the most relevant geometric references for the current localization hypothesis.

Given \(\mathcal{J}_t\) and the query descriptor \(d_t^q\), the procedure selects a compact reference sample set
\begin{equation}
  \mathcal{N}_t
  \subseteq
  \bigcup_{j\in\mathcal{J}_t}\mathcal{R}_j,
  \qquad
  |\mathcal{N}_t|\leq K,
  \label{eq:selected_references}
\end{equation}
where \(K\) denotes the maximum number of reference images. The reference whose descriptor is closest to the query is selected first. The remaining references are then chosen according to the descriptor-distance bounds \(d_{\min}\) and \(d_{\max}\), until either \(K\) references have been selected or no additional valid candidate remains. The complete procedure is summarized in \Cref{alg:reference_selection}.
This belief-guided selection substantially reduces the computational cost of FF3D inference while preserving the most relevant geometric references for the current localization hypothesis.

\begin{algorithm}[!htbp]
    \caption{Cluster-conditioned reference sample selection}
    \label{alg:reference_selection}
    \textbf{Input:} candidate place set $\mathcal{J}_t$, representative sample sets $\{\mathcal{R}_j\}_{j\in\mathcal{J}_t}$, query descriptor $d_t^q$, requested number of references $K$, and distance bounds $d_{\min}$ and $d_{\max}$.\\
    \textbf{Output:} selected reference sample set $\mathcal{N}_t\subseteq\bigcup_{j\in\mathcal{J}_t}\mathcal{R}_j$.
    \begin{pseudocode}
        \pstate Collect the samples in $\{\mathcal{R}_j\}_{j\in\mathcal{J}_t}$ into an ordered candidate list.
        \pstate L2-normalize the candidate descriptors and the query descriptor $d_t^q$.
        \pstate Select the candidate sample with the highest descriptor similarity to $d_t^q$ and use it to initialize $\mathcal{N}_t$.
        \pstate \textbf{while} $|\mathcal{N}_t|<K$ \textbf{do}
        \pstate \pindent For each unselected candidate sample $m_i$, compute
        \[
        \delta_i=\min_{m_r\in\mathcal{N}_t}d_{\cos}(d_i,d_r).
        \]
        \pstate \pindent Define the valid candidate set
        \[
        \mathcal{V}=\left\{m_i:d_{\min}\leq\delta_i\leq d_{\max}\right\}.
        \]
        \pstate \pindent \textbf{if} $\mathcal{V}=\varnothing$ \textbf{then break}.
        \pstate \pindent Add to $\mathcal{N}_t$ the sample in $\mathcal{V}$ with the smallest value of $\delta_i$.
        \pstate \textbf{end while}
        \pstate \textbf{return} $\mathcal{N}_t$.
    \end{pseudocode}
\end{algorithm}

\paragraph{Metric pose estimation.}
\label{sec:description:metric_pose_estimation}
Given the selected reference sample set \(\mathcal{N}_t\), the FF3D model \(g\) jointly processes the query image and the corresponding reference images:
\begin{equation}
  \left\{\hat{T}_q^g\right\}
  \cup
  \left\{\hat{T}_i^g\right\}_{m_i\in\mathcal{N}_t}
  =
  g\!\left(
    \left\{I_t^q\right\}
    \cup
    \left\{I_i\right\}_{m_i\in\mathcal{N}_t}
  \right),
  \label{eq:metric_model_call}
\end{equation}
following the general formulation of \Cref{eq:ff3d}. The model predicts the camera transforms of both the query and the selected reference images in its own internal coordinate frame. Since the particle filter operates in planar space, only the planar translation and yaw components of each transform are retained, yielding the corresponding poses \(\hat{p}_k^g\in\mathrm{SE}(2)\).

Because FF3D predicts poses only up to an arbitrary similarity transformation, the estimated poses must be aligned with the global map reference frame before being incorporated into the particle filter. To this end, the predicted poses of the selected reference images, \(\{\hat{p}_i^g\}_{m_i\in\mathcal{N}_t}\), are matched with their known map poses, \(\{p_i\}_{m_i\in\mathcal{N}_t}\). A robust \(\mathrm{Sim}(2)\) transformation is estimated using Least Median of Squares~\citep{rousseeuw1984least} over Umeyama fits~\citep{umeyama1991least}, yielding the scale, rotation, and translation parameters \((s_t,R_t,\tau_t)\). Applying this transformation to the predicted query pose \(\hat{p}_q^g\) produces the metric pose estimate \(\hat{x}_t^g\in\mathrm{SE}(2)\) expressed in the map reference frame.

\paragraph{Belief conditioning.}
As a final remark, note that, unlike initialization, where $\mathcal{J}_0$ is selected uniquely from the query descriptor, the online candidate set is conditioned on the propagated particle distribution. Consequently, the clusters in $\mathcal{J}_t$ are not required to be topologically adjacent. Therefore, a concentrated belief typically produces a spatially coherent reference set, whereas a multimodal belief can preserve references from separated hypotheses.

\subsubsection{Particle Update.}
\label{sec:description:particle_update}

Particle weights are updated using two different likelihoods: metric and appearance.

\paragraph{Metric likelihood.} The metric likelihood scores the consistency between the planar metric observation $\hat{x}_t^g$ and each propagated particle:
\begin{equation}
  \ell_{\mathrm{met}}^{[n]} =
  \exp\!\left(
    -\frac{
      d_{\mathrm{SE}(2)}\!\left(\hat{x}_t^{g},\, x_t^{[n]}\right)^2
    }{2\sigma_g^2}
  \right),
  \label{eq:met_lik}
\end{equation}
where $d_{\mathrm{SE}(2)}$ adds the Euclidean translation error in meters and the wrapped absolute yaw error in radians with unit rotational weight, such that one radian has the same numerical contribution as one meter. This likelihood converts the FF3D estimate into a particle-wise observation term.

\paragraph{Appearance likelihood.} Following the observation model of~\cite{jaenal2022unsupervised}, the appearance likelihood of each propagated particle is computed from the descriptor distribution of its associated cluster:
\begin{equation}
  \ell_{\mathrm{app}}^{[n]} =
  P\!\left(d_t^q \mid C_{c_t^{[n]}}\right).
  \label{eq:app_lik}
\end{equation}
This probability is evaluated using the descriptor distribution parameters of cluster $c_t^{[n]}$ defined in \Cref{sec:description:cluster_model}.
Thus, the appearance term scores the compatibility between the query descriptor and the descriptor distribution of the topological place associated with the propagated particle.

\paragraph{Likelihood fusion.} The appearance and metric likelihoods are fused so that both terms contribute additively to the particle update. Since both likelihoods may have different numerical scales, we fuse them in log-space, denoted as $\lambda = \log \ell$. We first normalize their logarithms independently across the particle set as
\begin{equation}
  \tilde{\lambda}_{k}^{[n]} =
  \frac{
    \lambda_k^{[n]} - \mu_k^\lambda
  }{
    \sigma_k^\lambda
  },
  \label{eq:norm_liks}
\end{equation}
where \(k\in\{\mathrm{app},\mathrm{met}\}\), and \(\mu_k^\lambda\) and \(\sigma_k^\lambda\) are respectively the mean and standard deviation of \(\lambda_k^{[n]}\) over the particle set. This normalization makes the appearance and metric scores comparable before applying the mixing parameter \(\alpha\):
\begin{equation}
  \lambda_{\mathrm{fuse}}^{[n]} =
  (1-\alpha)\,\tilde{\lambda}_{\mathrm{app}}^{[n]}
  + \alpha\,\tilde{\lambda}_{\mathrm{met}}^{[n]} .
  \label{eq:fused_log_lik}
\end{equation}
Setting \(\alpha=0\) recovers the appearance-only particle update, whereas setting \(\alpha=1\) weights particles only by the metric observation. The weights are then updated as
\begin{equation}
  \tilde{w}_t^{[n]} =
  {w}_{t-1}^{[n]}
  \exp\!\left(\lambda_{\mathrm{fuse}}^{[n]}\right),
  \label{eq:weight_update}
\end{equation}
followed by normalization,
\begin{equation}
  w_t^{[n]} =
  \frac{\tilde{w}_t^{[n]}}
  {\sum_m \tilde{w}_t^{[m]}}.
  \label{eq:weight_normalization}
\end{equation}

\subsubsection{Pose Estimation and Particle Resampling.}
\label{sec:description:particle_resampling}

The final camera pose estimate $\hat{x}_t$ is obtained by first identifying the dominant mode of the particle distribution and then computing a weighted $\mathrm{SE}(2)$ mean over the particles belonging to that mode.

To prevent particle degeneracy, the effective sample size is subsequently computed:
\begin{equation}
  N_{\mathrm{eff}} =
  \frac{1}{\sum_n \left(w_t^{[n]}\right)^2}.
  \label{eq:effective_sample_size}
\end{equation}
Whenever $N_{\mathrm{eff}}$ falls below a fixed fraction of the total number of particles $N_p$, the particle set is resampled.

% sections/04_experiments.tex
\section{EXPERIMENTS}
\label{sec:experiments}

We evaluate the proposed system on three publicly available datasets that cover indoor and outdoor localization under changes in illumination, weather, and viewpoint. Within each dataset, all evaluated methods use the same georeferenced map, query sequences, and post-initialization evaluation frames.
We first compare the complete system with two principal baselines in terms of localization accuracy, robustness to large errors, and deployable map size.
We then examine the interchangeable visual and geometric components, particle count, likelihood fusion, and map compression.
Finally, a case study examines whether the sequential topometric belief can remain localized when perceptual aliasing causes direct retrieval to select spatially inconsistent references.

\subsection{Datasets}
\label{sec:experiments:datasets}

% \FloatBarrier
\begin{figure*}[!ht]
\centering
\begin{subfigure}[t]{0.31\textwidth}
\centering
\includegraphics[width=\linewidth]{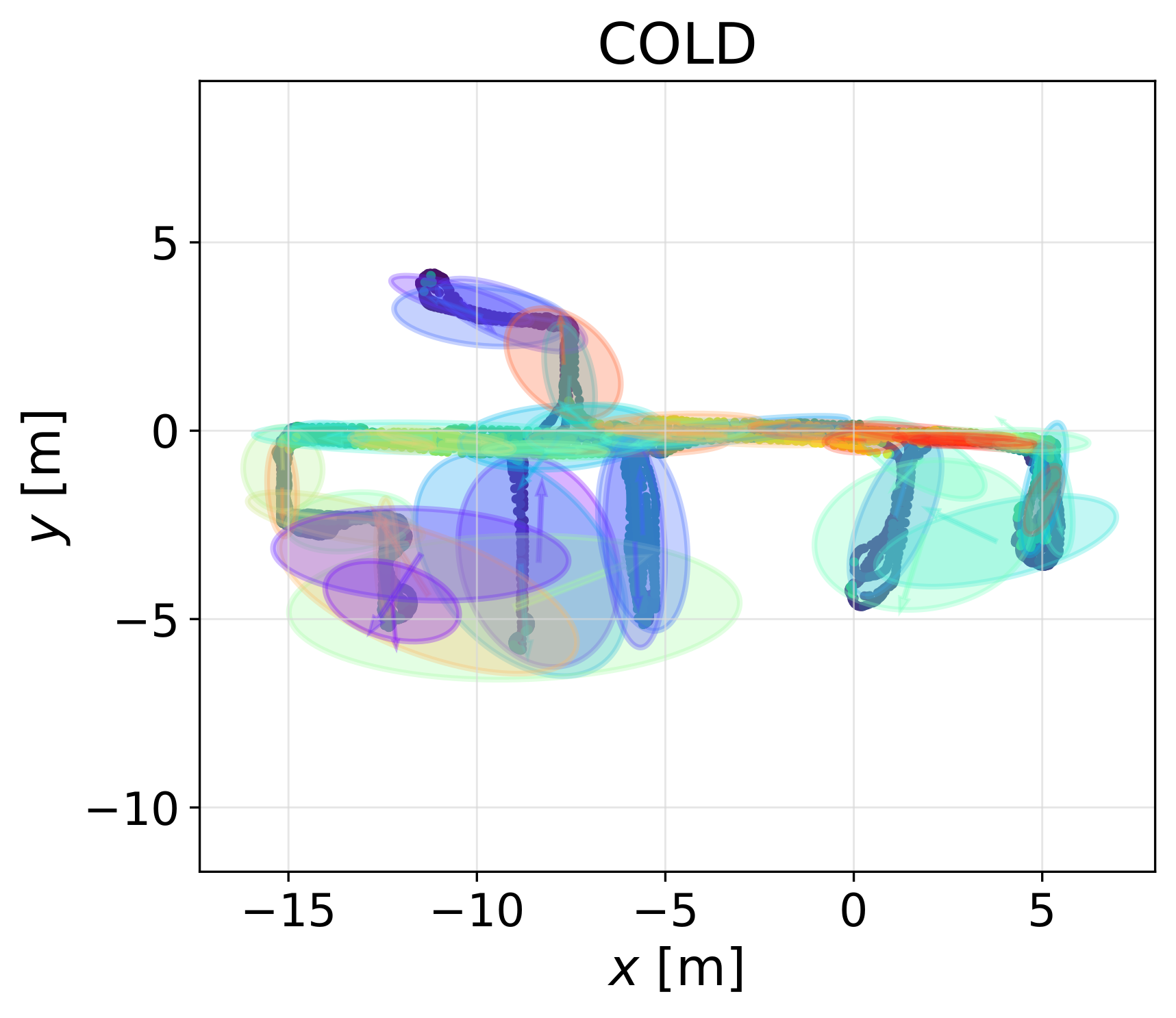}
% \caption{COLD.}
\label{fig:map_places_cold}
\end{subfigure}
\hfill
\begin{subfigure}[t]{0.31\textwidth}
\centering
\includegraphics[width=\linewidth]{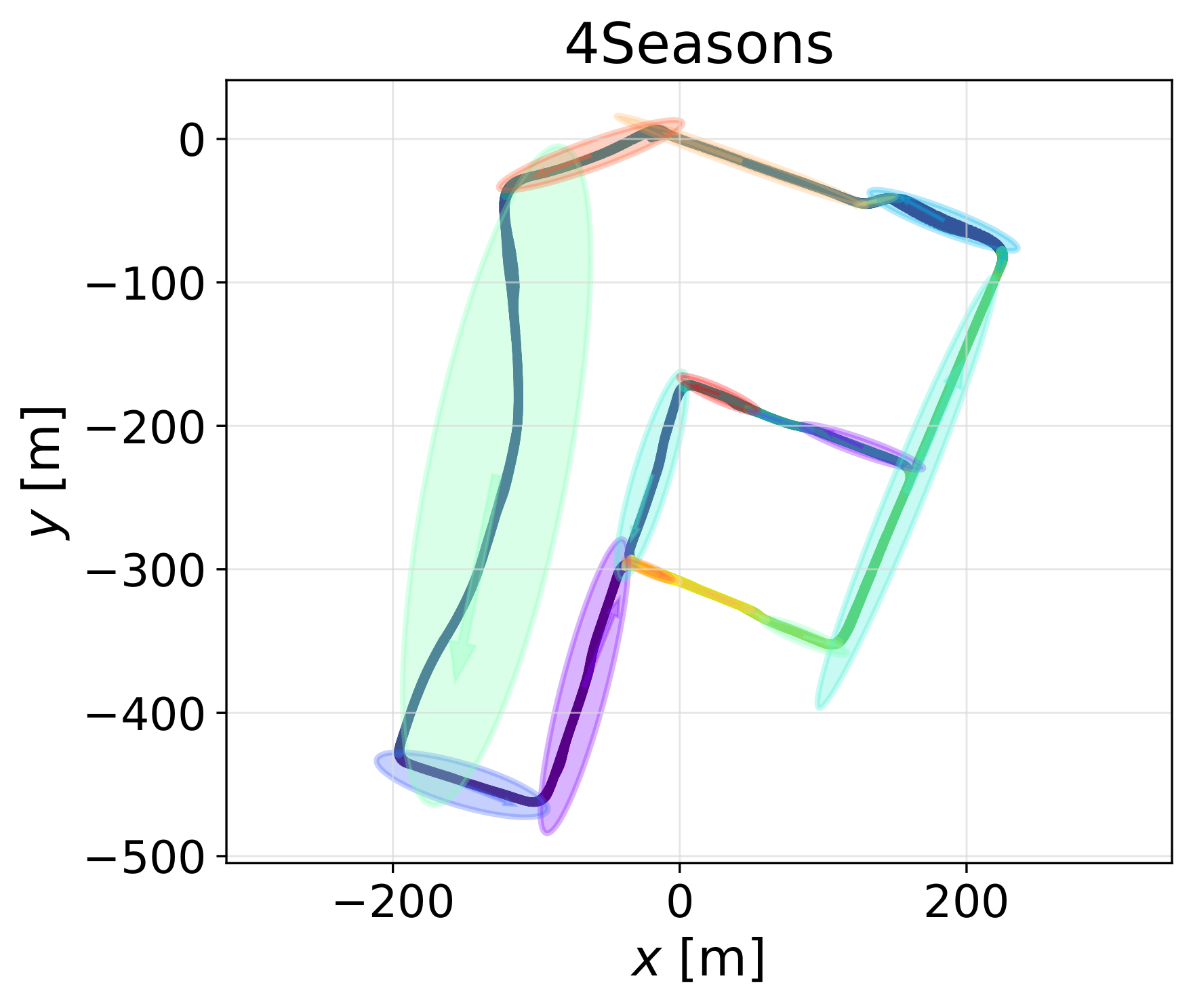}
% \caption{4Seasons.}
\label{fig:map_places_four_seasons}
\end{subfigure}
\hfill
\begin{subfigure}[t]{0.31\textwidth}
\centering
\includegraphics[width=\linewidth]{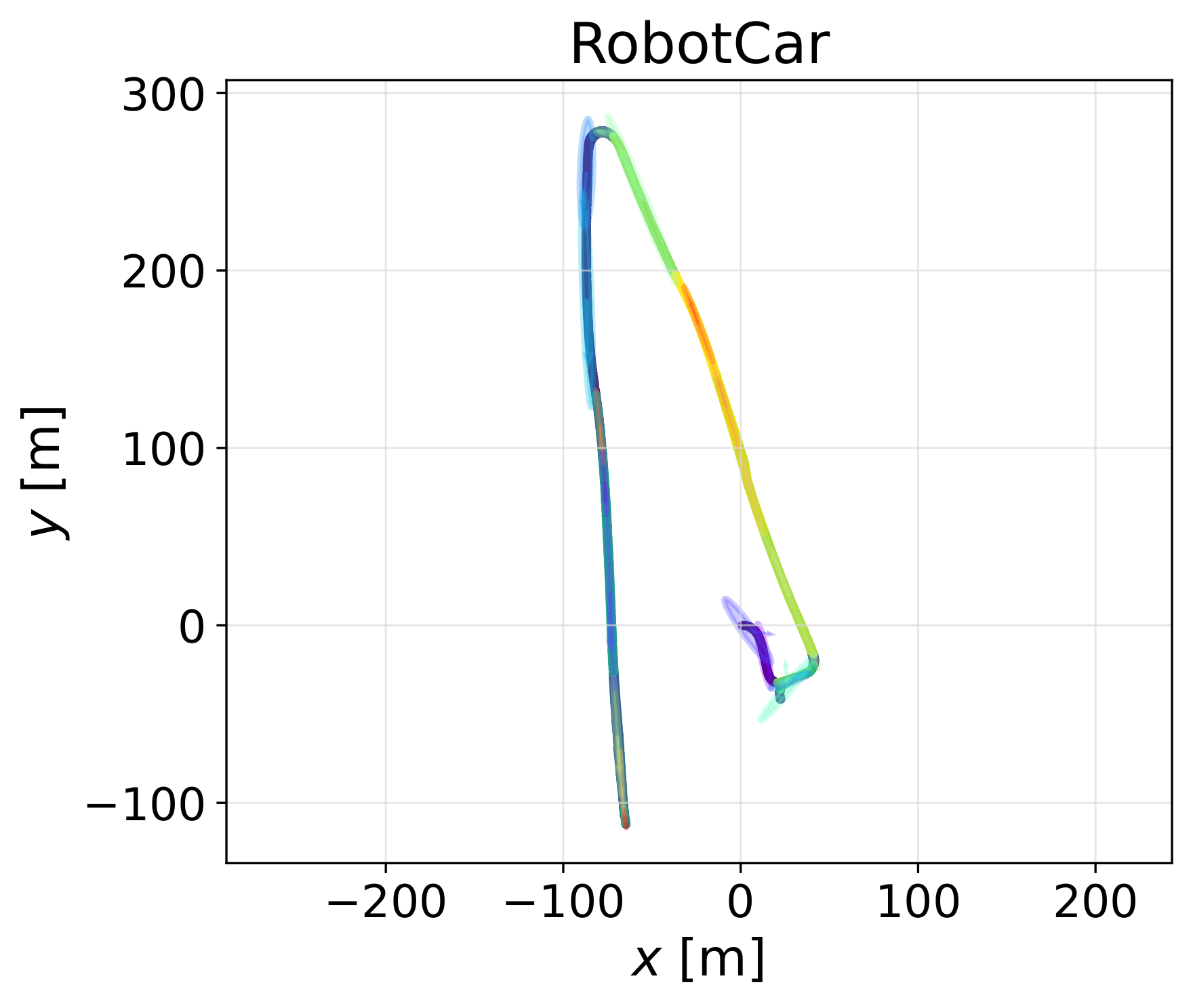}
% \caption{RobotCar.}
\label{fig:map_places_robotcar}
\end{subfigure}
\vspace{-14px}
\caption{Topometric maps constructed from the three reference sets. Retained reference samples are grouped by assigned place, ellipses show the planar place covariance, and arrows indicate the mean heading. Best in color.}
\vspace{-10px}
\label{fig:map_places}
\end{figure*}

% \FloatBarrier

% The selected map and query sequences for each dataset are described below.
We use the following datasets:

\textbf{COLD.} The COLD database~\citep{pronobis2009cold} contains indoor image sequences acquired by a mobile robot under different illumination conditions. We use the sequences recorded at the Autonomous Intelligent Systems Laboratory of the University of Freiburg. The georeferenced map combines the six cloudy sequences \textit{Seq1 cloudy1--3} and \textit{Seq2 cloudy1--3}, for a total of 12,482 images. The query set comprises \textit{Seq1 night1}, \textit{Seq2 night1}, and \textit{Seq1 sunny1}, thereby evaluating the same mapped environment under nighttime and sunny illumination.

\textbf{4Seasons.} The 4Seasons dataset~\citep{wenzel2020fourseasons} provides outdoor driving sequences with globally consistent poses. The map comprises \textit{Neighborhood} sequences \textit{1, 2}, and \textit{6}, with 8,620 images in total. We use \textit{Neighborhood} sequences \textit{3, 4} and \textit{7} as independent queries visiting the same area. We use undistorted images from the first camera.

\textbf{RobotCar.} The RobotCar dataset~\citep{maddern2017robotcar} provides repeated urban sequences recorded under varied weather conditions. The map uses sequences \textit{2015-03-03-11-31-36} and \textit{2015-07-29-13-09-26}, yielding 6,550 images. The query sequences are \textit{2015-08-12-15-04-18} (sun), \textit{2015-10-29-12-18-17} (rain), and \textit{2015-08-13-16-02-58} (overcast). We use the center camera and interpolated RTK poses~\citep{maddern2020rtk}. To keep repeated evaluation tractable, all traversals are restricted to a common route segment and every tenth query image is evaluated.

The first query image initializes the system and is excluded from the error statistics. The topometric maps constructed from the three reference sets are shown in \Cref{fig:map_places}. %\Cref{fig:map_places_cold,fig:map_places_four_seasons,fig:map_places_robotcar}.
HDBSCAN obtains 47, 17, and 23 places for COLD, 4Seasons, and RobotCar, respectively.

\subsection{Implementation Details}
\label{sec:experiments:implementation}

% VERSION TABLA ------------------------
The default feature-extraction, map-construction, filtering, and motion parameters, together with the hardware used, are summarized in \Cref{tab:experimental_parameters}. As part of the offline map-compression process, the representatives retained by FPS are resized to the input resolution required by \(g\) and stored as JPEG files at quality 95. Dataset odometry drives the motion model. Experiments were run on $3\times$ RTX 6000 Ada (48\,GiB).
The implementation scripts and evaluation benchmark will be made public upon acceptance.

Because all internal processes of particle-based methods are fundamentally stochastic, we execute each complete query sequence three times, using $\mathcal{S}=\{0,1,2\}$ as the random seeds. Each seed initializes the pseudorandom generator for one complete execution while the query images and all system parameters remain unchanged. 

% --------------------------------------
\paragraph{Baselines} We compare our method against two complete localization pipelines.
\textit{ALLOM} denotes the appearance-map localization framework developed in the series of papers~\citep{jaenal2022unsupervised,jaenal2023sequential}. Relying on K-Means+EM and ImRet descriptors, it served as the baseline for our work.
\textit{VPR+FF3D} retrieves \(K=9\) references from the complete map and applies DA3-Large for feed-forward 3D estimation. As a direct baseline, it uses the original map images without FPS, resizing, or JPEG recompression, and does not perform sequential filtering.

\begin{table}[!t]
\centering
\caption{Default experimental configuration.}
\label{tab:experimental_parameters}
\footnotesize
\begin{tabularx}{\columnwidth}{@{}>{\raggedright\arraybackslash}Xr@{}}
\toprule
\textbf{Parameter or component} & \textbf{Value or setting} \\
\midrule
\multicolumn{2}{@{}l}{\textbf{Models and storage}} \\
Descriptor extractor ($\phi$) & MixVPR-512~\citep{alibey2023mixvpr} \\
FF3D model ($g$) & DA3-Large~\citep{lin2025depthanything3} \\
Global descriptors & $\ell_2$-normalized \\
Map-image format & JPEG (quality 95, resized) \\
\midrule
\multicolumn{2}{@{}l}{\textbf{HDBSCAN and FPS sampling}} \\
Minimum cluster size / local density & 50 / 15 samples \\
Distance metric & Euclidean (\Cref{eq:pose_features}) \\
FPS diversity threshold ($\tau_{\mathrm{diversity}}$) & 0.075 \\
Representatives per place & Minimum: 2; maximum: $\infty$ \\
\midrule
\multicolumn{2}{@{}l}{\textbf{Filtering and reference selection}} \\
Particles ($N_p$) / mixing factor ($\beta$) & 200 / 0.5 \\
Likelihood params ($\alpha, \sigma_g$) & 0.70 / 0.20 \\
Ref. count ($K$) / dist. ($d_{\min},d_{\max}$) & 9 / $(0.10, 0.50)$ \\
Clustering / VPR thresh. ($\tau_{\mathrm{cl}}, \tau_{\mathrm{vpr}}$) & 0.10 / 0.80 \\
Resampling trigger & Multinomial ($N_{\mathrm{eff}} < N_p / 2$) \\
\midrule
\multicolumn{2}{@{}l}{\textbf{Motion-noise standard deviations $(x,y,\theta)$ [m, m, rad]}} \\
Indoor & $(0.025,0.010,0.015)$ \\
Outdoor & $(0.25,0.10,0.015)$ \\
\bottomrule
\end{tabularx}
\vspace{-15px}
\end{table}

\subsection{Evaluation Metrics}
\label{sec:experiments:metrics}

Let the ground-truth planar pose at time step $t$ be ${x_t^\text{gt}=(\mathbf{p}_t^\text{gt},\theta_t^\text{gt})}$ and let $\hat{x}_t=(\hat{\mathbf{p}}_t,\hat{\theta}_t)$ be the estimate obtained by our system. The position and yaw errors are
\begin{align}
    e_{p,t}      &= \left\|\hat{\mathbf{p}}_t-\mathbf{p}_t^\text{gt}\right\|_2,\\
    e_{\theta,t} &= \frac{180}{\pi}\,d_{\mathbb{S}^1}\!\left(\hat{\theta}_t,\theta_t^\text{gt}\right),
\end{align}
where $d_{\mathbb{S}^1}$ is the shortest angular distance on the circle.
We report the mean, median, and 90th percentile (P90) of these errors to characterize average accuracy, typical accuracy, and the error tail, respectively.

Nevertheless, these marginal statistics do not indicate how frequently the position and yaw requirements are satisfied simultaneously. Thus, let $\mathcal{T}$ be the set of evaluated time steps of one query sequence. The joint localization recall at position and yaw tolerances $(r,\delta)$ is given by
\begin{equation}
    R_{\mathrm{loc}}(r,\delta)=
    \frac{1}{|\mathcal{T}|}
    \sum_{t\in\mathcal{T}}
    \mathbf{1}\!\left[
    e_{p,t}\leq r \ \land\ e_{\theta,t}\leq\delta
    \right].
\label{eq:joint_localization_recall}
\end{equation}

Since a single tolerance pair gives the recall only at the selected limits and does not show how it changes under more demanding requirements, we scale both tolerances jointly and compute the normalized area
\begin{equation}
    \mathrm{AUC}(r,\delta)=
    \int_0^1
    R_{\mathrm{loc}}(\lambda r,\lambda\delta)\,
    \mathrm{d}\lambda,
\label{eq:joint_recall_auc}
\end{equation}
referred from now on as AUC. We use $(r,\delta)=(0.50~\mathrm{m},5^\circ)$ for COLD (indoor) and $(5~\mathrm{m},10^\circ)$ for 4Seasons and RobotCar (outdoor).

For particle-based methods, each metric is first computed over all evaluated frames of one complete run. The three repetitions of each query sequence are averaged, and the result per-dataset is obtained by averaging the values for the three query sequences with equal weight. The deterministic \textit{VPR+FF3D} baseline is executed once per query sequence, averaging the result per sequence to obtain the dataset result. We do not report standard deviations or make statistical-significance claims.
Some ablations report a single metric, denoted $\mathrm{AUC}_{\mathrm{avg}}$ obtained from averaging each dataset AUC with equal weight.

We define deployable map size as the total on-disk size of the map files required for online localization in MiB. For \textit{ALLOM}, this set contains only the serialized probabilistic map because reference images are not required online. For \textit{VPR+FF3D}, it contains the complete reference-image set and the associated map data, whereas for \textit{Ours} it contains the serialized topometric map and the resized JPEG representatives. Model checkpoints are excluded.

% \begin{equation}
%     \mathrm{AUC}_{\mathrm{avg}}
%     = \frac{1}{3}\sum_{D}\mathrm{AUC}_{D},
% \label{eq:auc_dataset_average}
% \end{equation}
% where the sum covers COLD, 4Seasons, and RobotCar.

% VERSION TEXTO ------------------------
%Unless stated otherwise, the descriptor extractor $\phi$ is MixVPR-512~\cite{alibey2023mixvpr} and the FF3D model $g$ is DA3-Large~\cite{lin2025depthanything3}. Global descriptors are $\ell_2$-normalized before map construction and online comparison. HDBSCAN uses a minimum cluster size of 50, a local-density parameter of 15 samples, and Euclidean distance over the standardized pose features in \Cref{eq:pose_features}. Samples initially labeled as noise are assigned to the nearest place centroid in the same standardized feature space.

%FPS uses $\tau_{\mathrm{diversity}}=0.075$, retains at least two representatives per place, and imposes no maximum. The retained map images are resized to the input size required by $g$ and stored as JPEG files at quality 95. The filter uses $N_p=200$ particles and $\beta=0.5$. Reference selection uses $K=9$, $(d_{\min},d_{\max})=(0.10,0.50)$, $\tau_{\mathrm{cl}}=0.10$, and $\tau_{\mathrm{vpr}}=0.80$. The likelihood parameters are $\alpha=0.70$ and $\sigma_g=0.20$.

% --------------------------------------

% LOCALIZATION ----------------- 

\begin{table*}[!t]
\centering
\caption{Overall localization results. Lower error and higher AUC are better; the best value in each dataset and column is bold. Underlined map sizes are the smallest among methods that retain reference images online.}
\label{tab:overall_localization}
\small
\begin{tabular}{@{}ll|rrr|rrr|r|r@{}}
\toprule
Dataset & Method & \multicolumn{3}{c|}{$e_p$ [m]} & \multicolumn{3}{c|}{$e_\theta$ [$^\circ$]} & \multirow{ 2}{*}{AUC} & Map size  \\
 & & Mean & Med.  & P90  & Mean & Med. & P90 & & [MiB] \\
\midrule
COLD
& \textit{ALLOM} & 0.313 & 0.259 & 0.576 & 4.670 & 4.070 & 9.680 & 0.192 & \textbf{30.7} \\
& \textit{VPR+FF3D} & 0.215 & 0.087 & 0.371 & 4.116 & \textbf{1.208} & 7.398 & 0.543 & 2164.6 \\
& \textit{Ours} & \textbf{0.124} & \textbf{0.081} & \textbf{0.232} & \textbf{2.104} & 1.455 & \textbf{4.912} & \textbf{0.553} & \underline{259.0} \\
\midrule
4Seasons
& \textit{ALLOM} & 56.313 & 36.981 & 112.519 & 22.587 & 16.703 & 39.719 & 0.047 & \textbf{12.3} \\
& \textit{VPR+FF3D} & 1.404 & \textbf{0.144} & 0.630 & \textbf{1.138} & \textbf{0.534} & \textbf{1.660} & \textbf{0.897} & 1210.7 \\
& \textit{Ours} & \textbf{0.268} & 0.173 & \textbf{0.558} & 1.228 & 0.770 & 2.852 & 0.864 & \underline{263.2} \\
\midrule
RobotCar
& \textit{ALLOM} & 4.946 & 4.686 & 8.593 & 3.837 & 2.542 & 10.234 & 0.201 & \textbf{11.1} \\
& \textit{VPR+FF3D} & 1.687 & \textbf{0.342} & 1.488 & 1.921 & \textbf{0.847} & 4.219 & 0.774 & 8818.8 \\
& \textit{Ours} & \textbf{0.494} & 0.411 & \textbf{1.027} & \textbf{1.749} & 1.282 & \textbf{3.891} & \textbf{0.798} & \underline{139.0} \\
\bottomrule
\end{tabular}
\vspace{-10px}
\end{table*}

\subsection{Localization Results}
\label{sec:experiments:overall}

This experiment determines how much the complete proposal improves over the appearance-based state of the art and whether sequential topometric reasoning offers advantages over direct retrieval followed by FF3D inference. 

\Cref{tab:overall_localization} compares localization accuracy and deployable map size. \textit{Ours} consistently improves upon \textit{ALLOM} and obtains the lowest mean and P90 position errors on all three datasets.
The lower mean and P90 errors show that the sequential belief limits the large position deviations produced when retrieval selects spatially inconsistent references. This is important for a continuously operating robot because an isolated estimate in the wrong map region can disrupt tracking or cause downstream navigation to act on an incorrect pose, even when most estimates are accurate. \textit{Ours} also achieves the highest AUC on COLD and RobotCar. On 4Seasons, direct \textit{VPR+FF3D} obtains lower median and yaw errors and a higher AUC, while \textit{Ours} retains lower mean and P90 position errors. Direct retrieval can therefore be highly precise when its references are correct, whereas the temporal belief provides greater protection against large spatial localization failures.
These accuracy gains are accompanied by a substantial reduction in deployable map size. The representative maps used by our method are considerably smaller than the complete image maps required by \textit{VPR+FF3D}. \textit{ALLOM} remains the smallest because it does not retain reference images for metric inference, but it also exhibits markedly lower localization accuracy. Overall, the proposed system provides a favorable compromise between compact map storage, sequential consistency, and localization accuracy across the three datasets and their appearance conditions.

% ABLATION ----------------- 
\subsection{Ablation Studies}
\label{sec:experiments:ablations}

The following experiments vary one subsystem at a time while keeping the remaining parameters at the default values in \Cref{sec:experiments:implementation}. %For each dataset, the results are aggregated over repetitions and query sequences using \Cref{eq:experiment_aggregation}. Figures or tables that summarize all three datasets report $\mathrm{AUC}_{\mathrm{avg}}$ from \Cref{eq:auc_dataset_average}. 
We use AUC as the primary selection metric because it jointly reflects position and yaw accuracy over increasingly demanding tolerances.

\subsubsection{Visual Descriptor.}
\label{sec:experiments:descriptor}

The global descriptor affects place likelihoods, reference retrieval, and descriptor storage. We therefore evaluate 19 configurations spanning NetVLAD~\citep{arandjelovic2016netvlad}, ImRet~\citep{radenovic2019imret}, CosPlace~\citep{berton2022cosplace}, EigenPlaces~\citep{berton2023eigenplaces}, MixVPR~\citep{alibey2023mixvpr}, AnyLoc~\citep{keetha2023anyloc}, CricaVPR~\citep{lu2024cricavpr}, MegaLoc~\citep{berton2025megaloc}, BoQ~\citep{ali2024boq}, and SALAD~\citep{izquierdo2024salad}. CosPlace is evaluated at five output sizes ranging from 128 to 2,048 dimensions, EigenPlaces at four sizes over the same range, and MixVPR at three sizes ranging from 128 to 4,096 dimensions. The remaining descriptor families are evaluated at their standard output dimensionality.

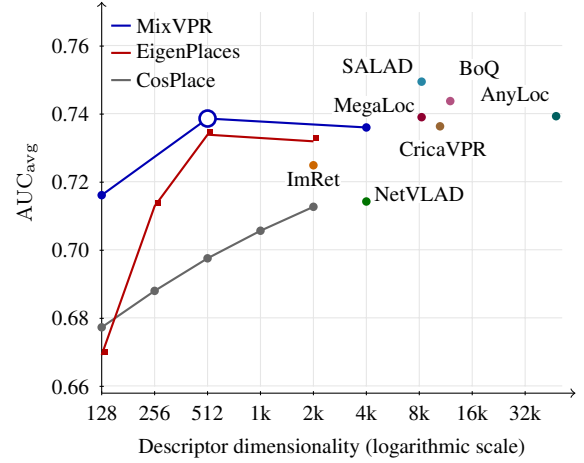
\begin{figure}[!t]
\centering
\begin{tikzpicture}[x=0.70cm,y=45cm,font=\footnotesize]
  \draw[gray!20] (0,0.66) grid[xstep=1,ystep=0.02] (8.7,0.77);
  \draw[->] (0,0.658) -- (8.95,0.658);
  \draw[->] (0,0.658) -- (0,0.773);

  \foreach \x/\lab in {
    0/128,1/256,2/512,3/1k,4/2k,
    5/4k,6/8k,7/16k,8/32k}
    \draw (\x,0.6574) -- (\x,0.6586)
      node[below=2pt] {\lab};

  \foreach \y/\lab in {
    0.66/0.66,0.68/0.68,0.70/0.70,0.72/0.72,
    0.74/0.74,0.76/0.76}
    \draw (-0.04,\y) -- (0.04,\y)
      node[left=2pt] {\lab};

  \node at (4.35,0.642)
    {Descriptor dimensionality (logarithmic scale)};
  \node[rotate=90] at (-1.40,0.72)
    {$\mathrm{AUC}_{\mathrm{avg}}$};

  % CosPlace variants
  \draw[black!60,thick] plot coordinates {
    (0,0.677354)
    (1,0.688057)
    (2,0.697629)
    (3,0.705728)
    (4,0.712774)
  };
  \foreach \x/\y in {
    0/0.677354,
    1/0.688057,
    2/0.697629,
    3/0.705728,
    4/0.712774}
    \fill[black!60] (\x,\y) circle[radius=1.6pt];

  % EigenPlaces variants
  \draw[red!70!black,thick] plot coordinates {
    (0,0.669229)
    (1,0.712938)
    (2,0.733968)
    (4,0.732017)
  };
  \foreach \x/\y in {
    0/0.669229,
    1/0.712938,
    2/0.733968,
    4/0.732017}
    \fill[red!70!black]
      (\x,\y) rectangle ++(2.2pt,2.2pt);

  % MixVPR variants
  \draw[blue!70!black,thick] plot coordinates {
    (0,0.716184)
    (2,0.738716)
    (5,0.736086)
  };
  \fill[blue!70!black]
    (0,0.716184) circle[radius=1.6pt];
  \filldraw[
    blue!70!black,
    fill=white,
    line width=1pt]
    (2,0.738716) circle[radius=3pt];
  \fill[blue!70!black]
    (5,0.736086) circle[radius=1.6pt];

  % Families evaluated at multiple descriptor dimensions
  \draw[blue!70!black,thick]
    (0.18,0.766) -- (0.55,0.766);
  \node[
    anchor=west,
    inner sep=0.5pt,
    text=black]
    at (0.62,0.766) {MixVPR};

  \draw[red!70!black,thick]
    (0.18,0.758) -- (0.55,0.758);
  \node[
    anchor=west,
    inner sep=0.5pt,
    text=black]
    at (0.62,0.758) {EigenPlaces};

  \draw[black!60,thick]
    (0.18,0.750) -- (0.55,0.750);
  \node[
    anchor=west,
    inner sep=0.5pt,
    text=black]
    at (0.62,0.750) {CosPlace};

  % Single-size descriptor families
  \fill[green!45!black]
    (5,0.714318) circle[radius=1.6pt];
  \node[
    anchor=west,
    fill=white,
    inner sep=0.5pt]
    at (5.12,0.7165) {NetVLAD};

  \fill[orange!80!black]
    (4,0.724981) circle[radius=1.6pt];
  \node[
    anchor=north,
    fill=white,
    inner sep=0.5pt]
    at (4,0.7235) {ImRet};

  \fill[purple!75!black]
    (6.044,0.739103) circle[radius=1.6pt];
  \node[
    anchor=east,
    fill=white,
    inner sep=0.5pt]
    at (5.95,0.742) {MegaLoc};

  \fill[cyan!60!black]
    (6.044,0.749541) circle[radius=1.6pt];
  \node[
    anchor=south east,
    fill=white,
    inner sep=0.5pt]
    at (5.92,0.7515) {SALAD};

  \fill[brown!80!black]
    (6.392,0.736419) circle[radius=1.6pt];
  \node[
    anchor=north,
    fill=white,
    inner sep=0.5pt]
    at (6.45,0.7325) {CricaVPR};

  \fill[magenta!70!black]
    (6.585,0.743835) circle[radius=1.6pt];
  \node[
    anchor=south west,
    fill=white,
    inner sep=0.5pt]
    at (6.72,0.75) {BoQ};

  \fill[teal!75!black]
    (8.585,0.739399) circle[radius=1.6pt];
  \node[
    anchor=south east,
    fill=white,
    inner sep=0.5pt]
    at (8.50,0.742) {AnyLoc};
\end{tikzpicture}
\vspace{-5px}
\caption{Visual-descriptor ablation. Lines connect dimensional variants of the same descriptor family, while individual markers denote architectures evaluated at one dimensionality. The enlarged marker identifies the selected configuration.}
\label{fig:descriptor_tradeoff}
\vspace{-10px}
\end{figure}

\Cref{fig:descriptor_tradeoff} compares $\mathrm{AUC}_{\mathrm{avg}}$ with descriptor dimensionality, with variants of the same architecture connected to expose how representation size affects localization.
Increasing dimensionality generally benefits the overall performance, although the improvement is not strictly monotonic. SALAD obtains the highest $\mathrm{AUC}_{\mathrm{avg}}$, but requires a substantially larger representation. MixVPR-512 reaches a similar accuracy range with only 512 components and does not improve when expanded to 4,096 components. We therefore retain MixVPR-512 as a balanced choice between localization accuracy and descriptor storage, rather than as an absolute winner. Since $\phi$ is interchangeable, applications with different storage or accuracy priorities can select another descriptor without changing the remaining pipeline.

\subsubsection{Feed-forward Geometry Model.}
\label{sec:experiments:geometry}

The FF3D model determines the quality and computational cost of the metric observation. We vary $g$ over DUSt3R~\citep{wang2024dust3r}, Fast3R~\citep{yang2025fast3r}, VGGT~\citep{wang2025vggt}, $\pi^3$ and $\pi^3$-X~\citep{wang2025pi3}, MapAnything~\citep{keetha2026mapanything}, ZipMap~\citep{jin2026zipmap}, and the DA3 family~\citep{lin2025depthanything3}.

\Cref{tab:geometry_ablation} compares $\mathrm{AUC}_{\mathrm{avg}}$ with FF3D backend latency, retaining DA3-Small, DA3-Large, and DA3-Giant-Rays as representative operating points from the DA3 family. Latency measures synchronized pose-backend computation for one query and $K=9$ references, excluding image loading and the remaining pipeline.

\begin{table}[!htbp]
\centering
\caption{FF3D-model ablation comparing backend latency and localization AUC. Lower latency and higher $\mathrm{AUC}_{\mathrm{avg}}$ are better; the best value in each column is bold.}
\label{tab:geometry_ablation}
\small
\begin{tabular}{@{}lrr@{}}
\toprule
FF3D model & Latency [s] & $\mathrm{AUC}_{\mathrm{avg}}$ \\
\midrule
DUSt3R & 7.523 & 0.674 \\
Fast3R & 4.988 & 0.714 \\
VGGT & 0.388 & 0.514 \\
$\pi^3$ & 0.427 & 0.740 \\
$\pi^3$-X & 0.495 & 0.728 \\
MapAnything & 0.447 & 0.715 \\
ZipMap & 0.549 & 0.689 \\
DA3-Small & \textbf{0.038} & 0.711 \\
DA3-Large & 0.237 & 0.739 \\
DA3-Giant-Rays & 0.576 & \textbf{0.743} \\
\bottomrule
\end{tabular}
\end{table}

The results show no monotonic relationship between computational cost and localization accuracy. DA3-Giant-Rays obtains the highest $\mathrm{AUC}_{\mathrm{avg}}$, but DA3-Large reaches a similar accuracy range with substantially lower latency. $\pi^3$ is also competitive but slower, whereas DA3-Small provides the lowest latency at the cost of reduced accuracy. We therefore retain DA3-Large as a balanced operating point. Since $g$ is interchangeable, another model can be selected when an application prioritizes either maximum accuracy or minimum inference time.

\subsubsection{Number of Particles.}
\label{sec:experiments:particles}

The particle count $N_p$ determines how densely the posterior over $\mathrm{SE}(2)$ can be represented. With too few particles, broad or multimodal beliefs may be approximated poorly and valid pose hypotheses can disappear during resampling; increasing $N_p$ improves belief coverage but raises the cost of motion propagation, likelihood evaluation, and resampling at every frame.

\begin{figure}[!htbp]
\centering
\begin{tikzpicture}[x=0.030cm,y=170cm,font=\footnotesize]
  \draw[gray!20] (0,0.718) grid[xstep=50,ystep=0.005] (190,0.745);
  \draw[->] (0,0.7175) -- (197,0.7175);
  \draw[->] (0,0.7175) -- (0,0.746);
  \foreach \x in {0,50,100,150}
    \draw (\x,0.7172) -- (\x,0.7178) node[below=2pt] {\x};
  \foreach \y/\lab in {0.720/0.720,0.725/0.725,0.730/0.730,0.735/0.735,0.740/0.740,0.745/0.745}
    \draw (-2.2,\y) -- (2.2,\y) node[left=2pt] {\lab};
  \node at (95,0.7140) {Particle-filter latency [ms/step]};
  \node[rotate=90] at (-39,0.7315) {$\mathrm{AUC}_{\mathrm{avg}}$};

  \draw[black!65,thick] plot coordinates {
    (5.60,0.71981) (10.98,0.73603) (26.38,0.73872)
    (50.21,0.74105) (104.29,0.74274) (185.08,0.74063)
  };
  \foreach \x/\y in {
    5.60/0.71981,10.98/0.73603,50.21/0.74105,
    104.29/0.74274,185.08/0.74063}
    \fill[black!65] (\x,\y) circle[radius=1.5pt];
  \filldraw[blue!70!black,fill=white,line width=1pt]
    (26.38,0.73872) circle[radius=3pt];

  \node[anchor=south west] at (6.8,0.7201) {50};
  \node[anchor=west] at (12.3,0.7361) {100};
  \node[anchor=south,blue!70!black] at (26.38,0.7393) {200};
  \node[anchor=south] at (50.21,0.7414) {500};
  \node[anchor=south] at (104.29,0.7431) {1000};
  \node[anchor=south east] at (183.5,0.7409) {2000};
\end{tikzpicture}
\caption{$\mathrm{AUC}_{\mathrm{avg}}$ and particle-filter latency as $N_p$ increases. The enlarged marker identifies the selected value $N_p=200$.}
\label{fig:particle_count_tradeoff}
\vspace{-10px}
\end{figure}
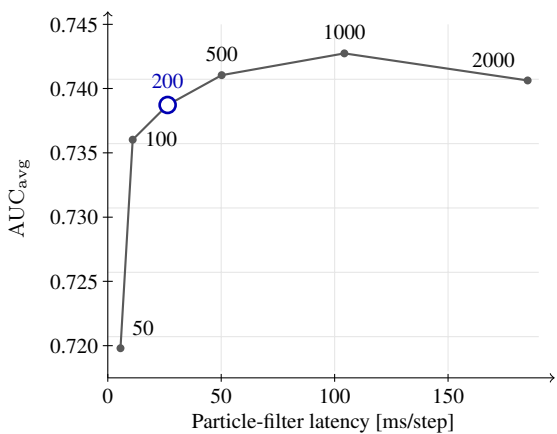

\Cref{fig:particle_count_tradeoff} compares $\mathrm{AUC}_{\mathrm{avg}}$ with the latency attributable exclusively to particle-filter computation, excluding reference selection and FF3D inference, as $N_p$ varies from 50 to 2,000. 
The largest improvement occurs before reaching 200 particles. Beyond this point, AUC varies only marginally, while the filtering cost continues to increase, indicating that the particle representation has entered a saturation region. Consequently, we select $N_p=200$ as a compromise that offers a reasonable balance between AUC and complexity.

\subsubsection{Fusion Weight.}
\label{sec:experiments:fusion}

The fusion parameter $\alpha$ determines whether the particle update can benefit from complementary appearance and metric evidence. $\alpha=0$ and $\alpha=1$ correspond to appearance-only and metric-only updates, respectively, whereas intermediate values combine both sources. 

\begin{figure}[H]
\centering
\begin{tikzpicture}[x=6.0cm,y=5.8cm,font=\footnotesize]
  \draw[gray!20] (0,0) grid[xstep=0.2,ystep=0.1] (1.0,0.77);
  \draw[->] (0,0) -- (1.04,0);
  \draw[->] (0,0) -- (0,0.79);
  \foreach \x/\lab in {0/0,0.2/0.2,0.4/0.4,0.6/0.6,0.8/0.8,1.0/1.0}
    \draw (\x,-0.012) -- (\x,0.012) node[below=2pt] {\lab};
  \foreach \y/\lab in {0.1/0.1,0.2/0.2,0.3/0.3,0.4/0.4,0.5/0.5,0.6/0.6,0.7/0.7}
    \draw (-0.012,\y) -- (0.012,\y) node[left=2pt] {\lab};
  \node at (0.5,-0.105) {Mixing parameter $\alpha$};
  \node[rotate=90] at (-0.16,0.4) {$\mathrm{AUC}_{\mathrm{avg}}$};

  \draw[black!65,thick] plot coordinates {
    (0.0,0.05233) (0.1,0.31018) (0.2,0.52537)
    (0.3,0.58908) (0.4,0.67102) (0.5,0.70424)
    (0.6,0.72571) (0.7,0.73872) (0.8,0.73716)
    (0.9,0.73263) (1.0,0.73517)
  };
  \foreach \x/\y in {
    0.0/0.05233,0.1/0.31018,0.2/0.52537,0.3/0.58908,
    0.4/0.67102,0.5/0.70424,0.6/0.72571,0.8/0.73716,
    0.9/0.73263,1.0/0.73517}
    \fill[black!65] (\x,\y) circle[radius=1.5pt];
  \filldraw[blue!70!black,fill=white,line width=1pt]
    (0.7,0.73872) circle[radius=3pt];
\end{tikzpicture}
\caption{$\mathrm{AUC}_{\mathrm{avg}}$ for different values of the parameter $\alpha$, which mixes appearance and metric sources. The highlighted marker identifies the selected value $\alpha=0.7$.}
\label{fig:fusion_weight_ablation}
\vspace{-10px}
\end{figure}
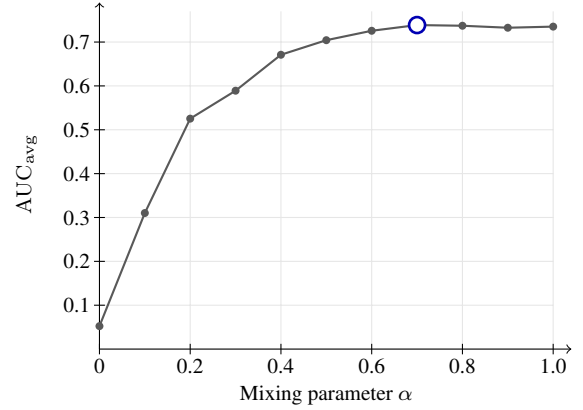

\Cref{fig:fusion_weight_ablation} reports the resulting $\mathrm{AUC}_{\mathrm{avg}}$. Appearance evidence alone is insufficient across the evaluated datasets, and performance rises sharply when metric information is introduced. The highest $\mathrm{AUC}_{\mathrm{avg}}$ occurs at the intermediate value $\alpha=0.7$, supporting the fusion of both observation terms. Larger values remain competitive but do not improve the three-dataset average, indicating that the appearance likelihood still provides complementary information.

\subsubsection{Map Compression.}
\label{sec:experiments:compression}

The threshold $\tau_{\mathrm{diversity}}$ controls the farthest-point sampling (FPS) stopping criterion: increasing it raises the minimum visual novelty required to retain another representative image and therefore produces a smaller map. This experiment determines how much visual redundancy can be removed before localization degrades. We evaluate all tested threshold values while keeping the encoding of retained images fixed. 

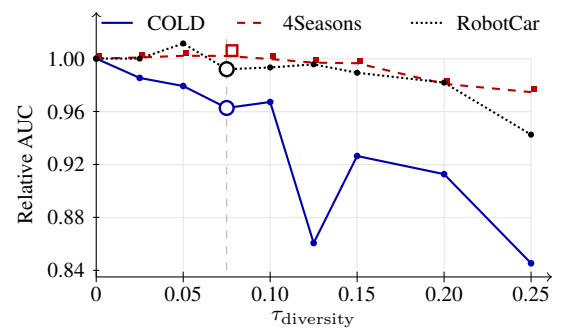
\begin{figure}[H]
\centering
\begin{tikzpicture}[x=23.0cm,y=17.5cm,font=\footnotesize]
  \draw[gray!20] (0,0.84) grid[xstep=0.05,ystep=0.04] (0.25,1.00);
  \draw[->] (0,0.835) -- (0.262,0.835);
  \draw[->] (0,0.835) -- (0,1.035);
  \foreach \x/\lab in {0/0,0.05/0.05,0.10/0.10,0.15/0.15,0.20/0.20,0.25/0.25}
    \draw (\x,0.831) -- (\x,0.839) node[below=2pt] {\lab};
  \foreach \y/\lab in {0.84/0.84,0.88/0.88,0.92/0.92,0.96/0.96,1.00/1.00}
    \draw (-0.0018,\y) -- (0.0018,\y) node[left=2pt] {\lab};
  \node at (0.125,0.803) {$\tau_{\mathrm{diversity}}$};
  \node[rotate=90] at (-0.041,0.92) {Relative AUC};
  \draw[gray!60,dashed] (0.075,0.84) -- (0.075,1.02);

  \draw[blue!70!black,thick] plot coordinates {
    (0,1.000000) (0.025,0.985383) (0.05,0.979260)
    (0.075,0.962690) (0.10,0.967189) (0.125,0.860632)
    (0.15,0.926378) (0.20,0.912673) (0.25,0.845319)
  };
  \draw[red!75!black,thick,dashed] plot coordinates {
    (0,1.000000) (0.025,1.000813) (0.05,1.002122)
    (0.075,1.001990) (0.10,0.999643) (0.125,0.996916)
    (0.15,0.996489) (0.20,0.980962) (0.25,0.974645)
  };
  \draw[black,thick,densely dotted] plot coordinates {
    (0,1.000000) (0.025,1.000017) (0.05,1.011392)
    (0.075,0.991980) (0.10,0.993278) (0.125,0.995631)
    (0.15,0.989289) (0.20,0.981840) (0.25,0.942537)
  };
  \foreach \x/\y in {
    0/1.000000,0.025/0.985383,0.05/0.979260,0.10/0.967189,
    0.125/0.860632,0.15/0.926378,0.20/0.912673,0.25/0.845319}
    \fill[blue!70!black] (\x,\y) circle[radius=1.25pt];
  \foreach \x/\y in {
    0/1.000000,0.025/1.000813,0.05/1.002122,0.10/0.999643,
    0.125/0.996916,0.15/0.996489,0.20/0.980962,0.25/0.974645}
    \fill[red!75!black] (\x,\y) rectangle ++(2.0pt,2.0pt);
  \foreach \x/\y in {
    0/1.000000,0.025/1.000017,0.05/1.011392,0.10/0.993278,
    0.125/0.995631,0.15/0.989289,0.20/0.981840,0.25/0.942537}
    \fill[black] (\x,\y) circle[radius=1.15pt];
  \filldraw[blue!70!black,fill=white,line width=0.9pt]
    (0.075,0.962690) circle[radius=2.6pt];
  \draw[red!75!black,line width=0.9pt]
    (0.075,1.001990) rectangle ++(4pt,4pt);
  \filldraw[black,fill=white,line width=0.9pt]
    (0.075,0.991980) circle[radius=2.6pt];

  \draw[blue!70!black,thick] (0.003,1.027) -- (0.021,1.027);
  \node[anchor=west] at (0.024,1.027) {COLD};
  \draw[red!75!black,thick,dashed] (0.081,1.027) -- (0.099,1.027);
  \node[anchor=west] at (0.102,1.027) {4Seasons};
  \draw[black,thick,densely dotted] (0.181,1.027) -- (0.199,1.027);
  \node[anchor=west] at (0.202,1.027) {RobotCar};
\end{tikzpicture}
\caption{AUC relative to the unreduced map as the FPS stopping threshold changes. Enlarged markers and the vertical line indicate the selected value $\tau_{\mathrm{diversity}}=0.075$.}
\label{fig:compression_tradeoff}
\vspace{-12px}
\end{figure}

\begin{figure*}[!htbp]
\centering
\begin{subfigure}[t]{0.31\textwidth}
\centering
{\footnotesize $(x,y,\theta)=(-5.73,-3.55,-92.77^\circ)$\par}
\smallskip
\includegraphics[width=\linewidth]{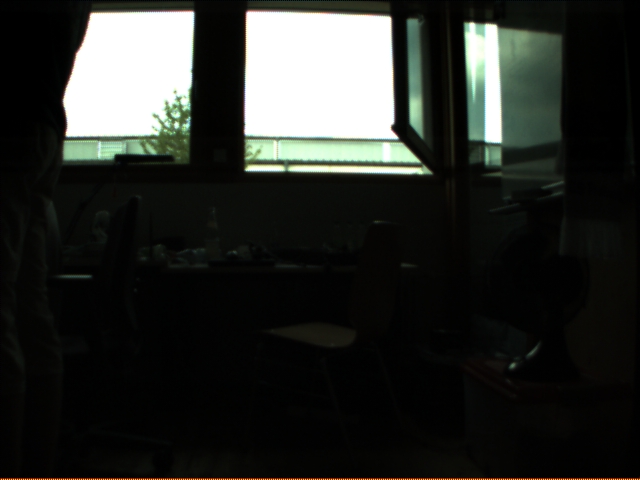}
\caption{Query.}
\label{fig:pa_cloudy_query}
\end{subfigure}
\hfill
\begin{subfigure}[t]{0.31\textwidth}
\centering
{\footnotesize $(x,y,\theta)=(-5.78,-2.91,-85.99^\circ)$\par}
\smallskip
\includegraphics[width=\linewidth]{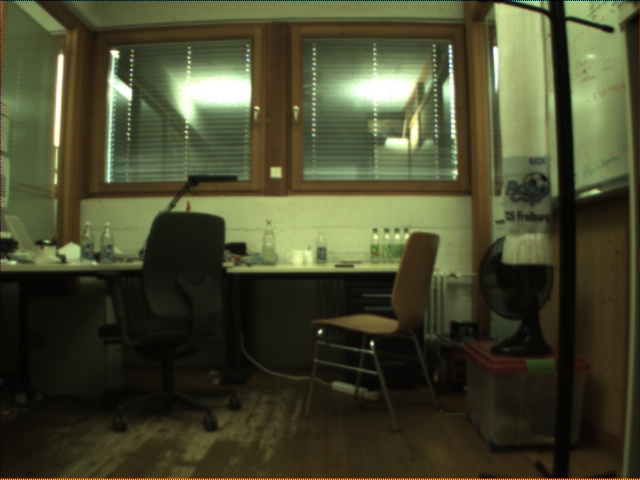}
\caption{\textit{Ours} reference.}
\label{fig:pa_cloudy_ours_reference}
\end{subfigure}
\hfill
\begin{subfigure}[t]{0.31\textwidth}
\centering
{\footnotesize $(x,y,\theta)=(1.16,-3.29,-93.85^\circ)$\par}
\smallskip
\includegraphics[width=\linewidth]{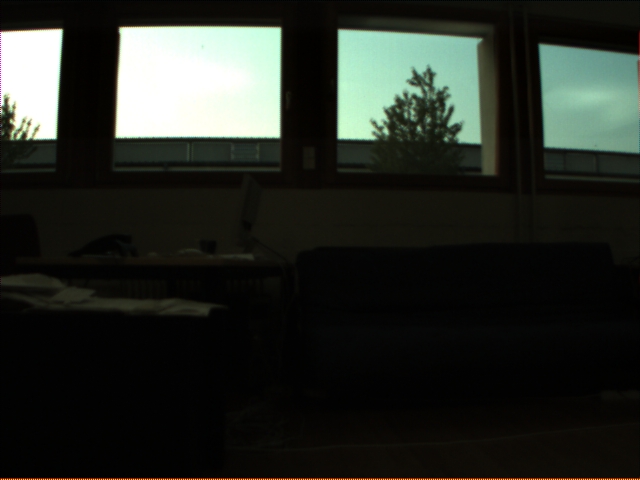}
\caption{\textit{VPR+FF3D} reference.}
\label{fig:pa_cloudy_vpr_reference}
\end{subfigure}
\caption{Reference selection at COLD frame 843. The values above each image give its ground-truth planar pose, with position in meters and yaw in degrees. Our method selects a reference in the tracked region, whereas direct \textit{VPR+FF3D} retrieves a visually similar image from an inconsistent location. The query and references are from the COLD database~\citep{pronobis2009cold}.}
\label{fig:pa_cloudy_qualitative}
\vspace{-12px}
\end{figure*}

\Cref{fig:compression_tradeoff} reports the ratio between the AUC at each threshold and the AUC obtained with the unreduced map. Localization accuracy does not vary monotonically with map reduction: while removing redundant views can occasionally improve reference selection, overly aggressive pruning risks discarding valuable visual context. Our default configuration ($\tau_{\mathrm{diversity}}=0.075$) maintains an AUC close to the unreduced baseline across all three datasets while achieving the compact map sizes reported in \Cref{tab:overall_localization}. Lower thresholds may be preferred when maximizing accuracy is the sole priority, whereas higher values (e.g., $0.15$) provide stronger compression for resource-constrained deployments. Thus, $\tau_{\mathrm{diversity}}=0.075$ represents a balanced cross-dataset operating point rather than a universal optimum.

% PERCEPTUAL ALIASING ------------------
\subsection{Robustness to Perceptual Aliasing}
\label{sec:experiments:pa}

Perceptual aliasing can make images acquired at different map locations appear similar, causing direct \textit{VPR+FF3D} to produce plausible poses in an incorrect map region. Although such episodes may occupy only a short portion of a query sequence, they are critical during continuous robot operation. This experiment therefore examines whether conditioning reference selection on the accumulated topometric belief prevents the metric estimate from following spatially inconsistent retrievals.

We use a COLD map built from \textit{Seq2 cloudy2} and localize the complete \textit{Seq2 cloudy1} query sequence. Although both sequences were recorded under cloudy conditions, different positions of the window blinds produce a substantial appearance change. The case was selected before running \textit{Ours}, using only ground truth and global VPR: we searched for consecutive frames for which a nearby mapped image existed but none of the \(K=9\) globally retrieved references belonged to the corresponding location. This procedure identified the episode and its temporal context shown over frames 820--900; the output of our method played no role in its selection.

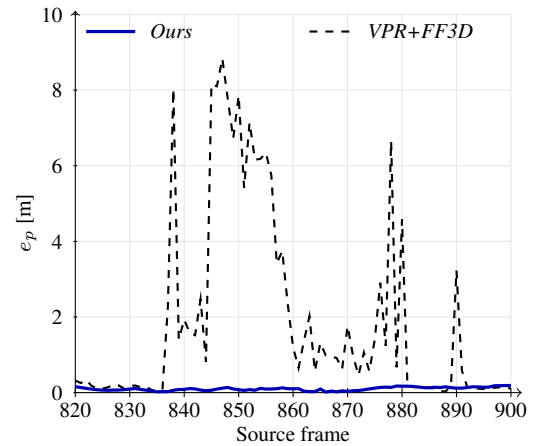
\begin{figure}[!htbp]
\centering
\begin{tikzpicture}[x=0.072cm,y=0.50cm,font=\footnotesize]
  \draw[gray!20] (0,0) grid[xstep=10,ystep=2] (80,10);
  \draw[->] (0,0) -- (82,0);
  \draw[->] (0,0) -- (0,10.1);
  \foreach \x/\lab in {0/820,10/830,20/840,30/850,40/860,50/870,60/880,70/890,80/900}
    \draw (\x,-0.12) -- (\x,0.12) node[below=2pt] {\lab};
  \foreach \y/\lab in {0/0,2/2,4/4,6/6,8/8,10/10}
    \draw (-0.35,\y) -- (0.35,\y) node[left=2pt] {\lab};
  \node at (40,-1.10) {Source frame};
  \node[rotate=90] at (-9.0,4.5) {$e_p$ [m]};

  \draw[black,thick,dashed] plot coordinates {
    (0,0.31965) (1,0.25287) (2,0.33591) (3,0.17762) (4,0.11541) (5,0.10703) (6,0.14209) (7,0.20566)
    (8,0.18896) (9,0.10062) (10,0.17323) (11,0.18855) (12,0.14095) (13,0.14832) (14,0.07373) (15,0.02854)
    (16,0.08221) (17,2.26796) (18,8.02609) (19,1.40031) (20,1.93045) (21,1.58442) (22,1.51865) (23,2.46696)
    (24,0.80933) (25,8.15551) (26,8.10206) (27,8.83450) (28,7.76689) (29,6.72376) (30,7.85559) (31,5.41551)
    (32,7.14178) (33,6.15874) (34,6.18248) (35,6.34399) (36,5.74661) (37,3.41320) (38,3.74801) (39,2.35694)
    (40,1.19620) (41,0.65122) (42,1.37632) (43,2.01622) (44,0.56614) (45,1.31212) (46,0.98481) (47,0.95798)
    (48,0.91479) (49,0.62219) (50,1.76206) (51,0.97678) (52,0.45822) (53,1.06170) (54,0.56496) (55,1.40834)
    (56,2.90537) (57,1.24099) (58,6.65306) (59,0.67133) (60,4.57922) (61,0.18387) (62,0.14688) (63,0.14433)
    (64,0.12809) (65,0.12233) (66,0.11277) (67,0.04813) (68,0.02408) (69,0.20097) (70,3.23041)
    (71,0.58169) (72,0.14374) (73,0.15578) (74,0.10451) (75,0.09761)
    (76,0.15281) (77,0.12473) (78,0.14285) (79,0.13660) (80,0.10707)
  };
  \draw[blue!70!black,very thick] plot coordinates {
    (0,0.15774) (1,0.13498) (2,0.11287) (3,0.09226) (4,0.07549) (5,0.06568) (6,0.06195) (7,0.06736)
    (8,0.06876) (9,0.07892) (10,0.09388) (11,0.10905) (12,0.08446) (13,0.06554) (14,0.04653) (15,0.02058)
    (16,0.02655) (17,0.02485) (18,0.06238) (19,0.08243) (20,0.08538) (21,0.10815) (22,0.09784) (23,0.06755)
    (24,0.05039) (25,0.06313) (26,0.09409) (27,0.11796) (28,0.13759) (29,0.09799) (30,0.07651) (31,0.05893)
    (32,0.08019) (33,0.06316) (34,0.11221) (35,0.09478) (36,0.09608) (37,0.10445) (38,0.12719) (39,0.10421)
    (40,0.10127) (41,0.10514) (42,0.03976) (43,0.02649) (44,0.02809) (45,0.09539) (46,0.01455) (47,0.04098)
    (48,0.02205) (49,0.04624) (50,0.03413) (51,0.05039) (52,0.05412) (53,0.07443) (54,0.09862) (55,0.11925)
    (56,0.13831) (57,0.14456) (58,0.13843) (59,0.17505) (60,0.16952) (61,0.17001) (62,0.16441) (63,0.14880)
    (64,0.13094) (65,0.12973) (66,0.14212) (67,0.13465) (68,0.15330) (69,0.13258) (70,0.11804)
    (71,0.11924) (72,0.13513) (73,0.15485) (74,0.15156) (75,0.14265)
    (76,0.14674) (77,0.18262) (78,0.18384) (79,0.18809) (80,0.18753)
  };
  \draw[blue!70!black,very thick] (2,9.55) -- (11,9.55);
  \node[anchor=west] at (12,9.55) {\textit{Ours}};
  \draw[black,thick,dashed] (43,9.55) -- (51,9.55);
  \node[anchor=west] at (52,9.55) {\textit{VPR+FF3D}};
\end{tikzpicture}
\caption{Position error during the selected COLD perceptual-aliasing case. Both methods process the complete query sequence from its beginning; the curve for \textit{Ours} is the mean across seeds.}
\label{fig:pa_cloudy_temporal}
\vspace{-15px}
\end{figure}

Both methods process the query sequence from its first image and are not initialized inside the selected episode. As shown in \Cref{fig:pa_cloudy_temporal}, direct \textit{VPR+FF3D} repeatedly loses the correct map region, whereas \textit{Ours} remains close to the ground truth throughout the same appearance ambiguity.

Direct \textit{VPR+FF3D} is attracted to a visually similar but spatially inconsistent part of the map, whereas the sequential estimate remains near the ground truth. Over the predefined interval, our method substantially reduces the median and P90 position errors and also lowers the median yaw error.

The qualitative example at frame 843 in \Cref{fig:pa_cloudy_qualitative} explains this behavior. Global retrieval selects an image that resembles the query but belongs to an inconsistent map location. In contrast, the belief-conditioned search restricts the candidate places to those supported by the preceding particle distribution and selects a reference from the tracked region. The sequential belief therefore changes where metric evidence is sought; it does not merely smooth an already computed FF3D trajectory. This focused experiment isolates the mechanism under perceptual aliasing and is not intended as a dataset-wide robustness score.

% sections/05_conclusion.tex

\section{CONCLUSION}
\label{sec:conclusions}

In this work, we have presented a modular topometric visual localization framework that combines appearance-robust VPR and FF3D metric pose estimation within a sequential probabilistic formulation. A dense georeferenced image map is abstracted into probabilistic places, while the particle belief preserves alternative pose hypotheses and conditions metric inference on spatially plausible references. Appearance and aligned FF3D likelihoods are fused in a common planar state space, and both the visual descriptor and FF3D model remain independently replaceable.

Experiments on COLD, 4Seasons, and RobotCar cover indoor and outdoor operation across pronounced illumination and weather changes. The proposed system substantially improves on the state-of-the-art appearance-based topometric methods and obtains the lowest mean and P90 position errors on every dataset. It also achieves the highest AUC on COLD and RobotCar while remaining competitive on 4Seasons. Direct VPR+FF3D can be highly accurate when retrieval identifies spatially consistent references, but its independent estimates provide no temporal mechanism for rejecting matches that conflict with the tracked trajectory. This distinction matters in mobile robotics, where a low typical error cannot compensate for occasional large failures. The abstracted scene representation also reduces deployable map storage by factors ranging from approximately five to more than sixty relative to dense VPR+FF3D maps.
The predefined perceptual-aliasing case study makes the role of the method explicit. Under a substantial appearance change, global retrieval repeatedly selects references from an inconsistent map location, whereas the topometric belief preserves the tracked mode and conditions metric estimation on spatially plausible places. The framework therefore does more than smooth FF3D outputs: it changes the evidence supplied to the metric estimator. 

However, the present formulation assumes a georeferenced map, odometry, and planar $\mathrm{SE}(2)$ motion, and its performance depends on the spatial consistency of the places and their references. Large or elongated places may admit inconsistent references and allow erroneous metric hypotheses to influence the belief, as observed on 4Seasons; while fixed likelihood fusion also does not explicitly represent FF3D confidence. Future work will investigate uncertainty-aware metric observations and adaptive fusion, place refinement based on spatial extent and belief uncertainty, and confidence-aware reference selection, together with extensions to full $\mathrm{SE}(3)$ motion and incrementally updated maps.

%%%%%%%%%%%%%%%%%%%%%%%%%%%%%%%%%%%%%%%%%%%%%%%%%%%%%%%%%%%%%%%%%%%%%%%%%%%%%%%%
% Notation
%%%%%%%%%%%%%%%%%%%%%%%%%%%%%%%%%%%%%%%%%%%%%%%%%%%%%%%%%%%%%%%%%%%%%%%%%%%%%%%%

% \input{sections/notes/notation}

%%%%%%%%%%%%%%%%%%%%%%%%%%%%%%%%%%%%%%%%%%%%%%%%%%%%%%%%%%%%%%%%%%%%%%%%%%%%%%%%
% Author contributions
%%%%%%%%%%%%%%%%%%%%%%%%%%%%%%%%%%%%%%%%%%%%%%%%%%%%%%%%%%%%%%%%%%%%%%%%%%%%%%%%

\section*{Author Contributions}

\noindent\hangindent=2em\hangafter=1\textbf{Eulogio Quemada-Torres:} Conceptualization, Methodology, Software, Validation, Formal analysis, Investigation, Data curation, Visualization, Writing -- original draft, Writing -- review \& editing.
\par\medskip

\noindent\hangindent=2em\hangafter=1\textbf{Alberto Jaenal:} Conceptualization, Methodology, Supervision, Writing -- review \& editing.
\par\medskip

\noindent\hangindent=2em\hangafter=1\textbf{Francisco-Angel Moreno:} Conceptualization, Supervision, Writing -- review \& editing.
\par\medskip

\noindent\hangindent=2em\hangafter=1\textbf{Javier Gonzalez-Jimenez:} Conceptualization, Supervision, Resources, Funding acquisition, Writing -- review \& editing.

%%%%%%%%%%%%%%%%%%%%%%%%%%%%%%%%%%%%%%%%%%%%%%%%%%%%%%%%%%%%%%%%%%%%%%%%%%%%%%%%
% Statements and Declarations
%%%%%%%%%%%%%%%%%%%%%%%%%%%%%%%%%%%%%%%%%%%%%%%%%%%%%%%%%%%%%%%%%%%%%%%%%%%%%%%%

\section*{Statements and Declarations}

\subsection*{Ethical considerations}

Ethical approval was not required because this study did not involve human or animal participants.

\subsection*{Consent to participate}

Not applicable.

\subsection*{Consent for publication}

Not applicable.

\subsection*{Declaration of conflicting interest}

The authors declared no potential conflicts of interest with
respect to the research, authorship, and/or publication of this
article.

\subsection*{Funding statement}

This work has been supported by the projects: MINDMAPS (PID2023-148191NB-I00), funded by the Ministry of Science, Innovation, and Universities of Spain and JDC2024-055088-I, funded by MICIU/AEI/10.13039/501100011033 and the FSE+.

\subsection*{Data availability}

The code, configurations, and scripts required to reproduce the reported results will be made publicly available upon acceptance. 
Third-party datasets and pretrained models are not redistributed and must be obtained from their original sources.

\subsection*{Use of AI tools}

During the preparation of this manuscript, the authors used OpenAI Codex (GPT-5) for English-language editing, LaTeX formatting, code inspection, and debugging. It was not used to generate experimental data, perform the reported experiments, or determine scientific conclusions. All AI-assisted outputs were critically reviewed and verified by the authors, including their technical correctness and any cited material where applicable. The authors acknowledge that AI-generated outputs may contain biases, errors, or omissions and take full responsibility for the accuracy, originality, and integrity of the manuscript.

% If supplementary material is included, uncomment:
%
% \begin{sm}
% Supplementary material for this article is available online.
% \end{sm}

%%%%%%%%%%%%%%%%%%%%%%%%%%%%%%%%%%%%%%%%%%%%%%%%%%%%%%%%%%%%%%%%%%%%%%%%%%%%%%%%
% References
%%%%%%%%%%%%%%%%%%%%%%%%%%%%%%%%%%%%%%%%%%%%%%%%%%%%%%%%%%%%%%%%%%%%%%%%%%%%%%%%

\bibliographystyle{SageH}
\bibliography{references}

\end{document}